\documentclass[preprint,onefignum,onetabnum]{siamonline190516}

\usepackage{amssymb, amsmath}
\usepackage{hyperref}

\usepackage{lipsum}
\usepackage{amsfonts}
\usepackage{graphicx}
\usepackage{epstopdf}
\usepackage{algorithmic}
\ifpdf
  \DeclareGraphicsExtensions{.eps,.pdf,.png,.jpg}
\else
  \DeclareGraphicsExtensions{.eps}
\fi

\usepackage{enumitem}
\setlist[enumerate]{leftmargin=.5in}
\setlist[itemize]{leftmargin=.5in}

\newsiamremark{remark}{Remark}
\newsiamremark{hypothesis}{Hypothesis}
\crefname{hypothesis}{Hypothesis}{Hypotheses}
\newsiamthm{claim}{Claim}

\headers{Weighted trigonometric interpolation}{Y. Xie, H. Chou, H. Rauhut, and R. Ward}

\title{
Overparameterization and generalization error: weighted trigonometric interpolation
\thanks{Submitted to the editors \today.
\funding{This work was funded by AFOSR 2018 MURI Award, DAAD grant 57417829 and Excellence Initiative of the German federal and state governments.}
}}

\author{Yuege Xie\thanks{Oden Institute, University of Texas at Austin, Austin TX 78712 USA (\email{yuege@oden.utexas.edu}).}
\and Hung-Hsu Chou\thanks{Chair for Mathematics of Information Processing, RWTH Aachen University, Pontdriesch 10, 52056 Aachen, Germany (\email{chou@mathc.rwth-aachen.de, rauhut@mathc.rwth-aachen.de}).} 
\and Holger Rauhut$^{\ddag}$ 
\and Rachel Ward$^{\dagger}$ \thanks{Mathematics Department, University of Texas at Austin, Austin TX 78712 USA (\email{rward@math.utexas.edu}).}}
\usepackage{amsopn}
\DeclareMathOperator{\diag}{diag}

\usepackage{amsmath,amsfonts,bm}

\def\eqref#1{equation~\ref{#1}}

\def\1{\bm{1}}

\def \diag{{\text{diag}}}
\def \st{{s.t.~}}

\def\v0{{\bm{0}}}

\def\vtheta{{\bm{\theta}}}
\def \vbeta{{\bm{\beta}}}
\def \vphi{{\bm{\phi}}}
\def \hbeta{{\hat{\bm{\beta}}}}

\def\vk{{\bm{k}}}

\def\vw{{\bm{w}}}
\def\vx{{\bm{x}}}
\def\vy{{\bm{y}}}

\def\mA{{\bm{A}}}
\def\mB{{\bm{B}}}
\def\mC{{\bm{C}}}

\def\mF{{\bm{F}}}

\def\mH{{\bm{H}}}
\def\mI{{\bm{I}}}

\def\mK{{\bm{K}}}
\def\mL{{\bm{L}}}
\def\mM{{\bm{M}}}
\def\mN{{\bm{N}}}
\def\mO{{\bm{O}}}

\def\mU{{\bm{U}}}

\DeclareMathAlphabet{\mathsfit}{\encodingdefault}{\sfdefault}{m}{sl}
\SetMathAlphabet{\mathsfit}{bold}{\encodingdefault}{\sfdefault}{bx}{n}

\def\tF{{\tens{F}}}

\def\gC{{\mathcal{C}}}

\def\gN{{\mathcal{N}}}

\def\gP{{\mathcal{P}}}
\def\gQ{{\mathcal{Q}}}

\def\sC{{\mathbb{C}}}

\def\sN{{\mathbb{N}}}

\def\sP{{\mathbb{P}}}

\def\sR{{\mathbb{R}}}

\def\sT{{\mathbb{T}}}

\def \ex{{\mathbb{E}}}

\def \tF {\Tilde{\mF}}
\def \bt {\vbeta_T}
\def \btc {\vbeta_{T^c}}

\def \Fst {{\mF_{T}}}

\def \Fstc {{\mF_{T^c}}}

\def \tFst {\tF_{T}}

\def \tFstc {{\tF_{T^c}}}

\def \dtFst {{ \tFst^{\dagger} }}

\def \sT {\Sigma_T}
\def \sTc {\Sigma_{T^c}}
\def \sTq {\sT^q}

\def \stu {\sT^u}
\def \stcu {\sTc^u}
\def \sTqs {\sT^{2q}}

\def \sTc {\Sigma_{T^c}}
\def \sTcr {\sTc^{2r}}
\def \sTcq {\sTc^q}

\def \kt {\mK_T}
\def \ktc {\mK_{T^c}}

\def \Sigmaq {\Sigma^q}

\def \mau {\mA_u}

\def \mcu {\mC_u}
\def \a2 {\mA_2}
\def \qa2 {\mA_{2q}}
\def \b2 {\mB_2}
\def \qb2 {\mB_{2q}}
\def \c2 {\mC_2}
\def \qc2 {\mC_{2q}}

\def \tjr {t_j^{2r}}

\def \un {\mU_n}
\def \fn {\mF_n}
\def \omn {\omega_n}

\def \smqs#1 { \sum_{k=0}^{p-1} \sum_{r=\lceil -k/n\rceil }^{\lfloor (p-1-k)/n \rfloor} t_k^{2q} t_{n r+k}^#1 e_{s,k} }

\def \tr#1 {\text{tr}\left(#1\right)}

\def \p#1 {{\left(#1\right)}}
\def \norms#1 {\left\|#1\right\|^2}
\def \bs#1 {{\left[#1\right]}}
\def \br#1 {{\left\{#1\right\}}}
\def \inp#1 {{\langle#1\rangle}}

\def \suma#1 { \sum_{k=0}^{n-1} \p{ \sum_{\nu=0}^{l-1}  t_{k+n\nu}^#1 } }
\def \sumc#1 {  \sum_{k=0}^{n-1} \p{\sum_{\nu =l}^{\tau -1} t_{k+n\nu}^#1 } }
\def \las#1 { \suma#1 \p{ \sum_{j=0}^{n-1}  \omn^{(s-k)j}} }
\def \lcs#1 { \sumc#1 \p{ \sum_{j=0}^{n-1}  \omn^{(s-k) j}} }
\def \esk {e_{s,k}}

\def \sinsuma#1 { \sum_{\nu=0}^{l-1}  t_{k+n\nu}^#1  }
\def \sinsumc#1 { \sum_{\nu =l}^{\tau  -1} t_{k+n\nu}^#1 } 

\def \sumr {  \sum_{j=0}^{D-1} t_j^{2r} }

\def \covr {c_r\Sigma^{2r}}
\def \crs {c_r}

\def \esk {e_{s,k}^{(n)}}

\def \aj {\sum_{\nu=0}^{l-1} \sum_{k=0}^{n-1} t_{k+n\nu}^u \omn^{-j k}}
\def \cj { \sum_{\nu=l}^{\tau -1}\sum_{k=0}^{n-1} t_{k+n\nu}^u \omn^{-j k} }

\def \a {\alpha}

\def \risk {\operatorname{risk}}

\def \htheta {\hat{\vtheta}}

\ifpdf
\hypersetup{
  pdftitle={Overparameterization and generalization error: weighted trigonometric interpolation},
  pdfauthor={}
}
\fi

\begin{document}

\maketitle

\begin{abstract}
Motivated by surprisingly good generalization properties of learned deep neural networks in overparameterized scenarios and by the related double descent phenomenon, this paper analyzes the relation between smoothness and low generalization error in an overparameterized linear learning problem. We study a random Fourier series model, where the task is to estimate the unknown Fourier coefficients from equidistant samples. We derive exact expressions for the generalization error of both plain and weighted least squares estimators. We show precisely how a bias towards smooth interpolants, in the form of weighted trigonometric interpolation, can lead to smaller generalization error in the overparameterized regime compared to the underparameterized regime. This provides insight into the power of overparameterization, which is common in modern machine learning.
\end{abstract}



\section{Introduction}
Consider the regression/interpolation problem: Given training data $({\bf x}_j, y_j) \in {\cal D} \times \mathbb{C}$, $j=1,\hdots,n$, corresponding to samples of an unknown function $y = f({\bf x})$ and the sampling points drawn from ${\cal D} \subset \mathbb{R}^d$, we would like to fit the data to a hypothesis class $\mathcal{H}:=\{f_\vtheta({\bf x}): \mathbb{R}^d\rightarrow\mathbb{C}, \vtheta \in \mathbb{C}^p\} $ by solving for parameters minimizing the empirical $\ell_2$-risk 
\begin{align}\label{eq:ls}
    \vtheta_{opt} \in \underset{ \vtheta \in \mathbb{C}^p}{\arg \min}  \sum_{j=1}^n | f_\vtheta(\vx_j) - y_j|^2.
\end{align}

Traditionally, the number of parameters $p$ is restricted to be smaller than the number of training samples, i.e., $p \leq n$, to avoid overfitting. For $p >n$, the solution $\vtheta_{opt}$ is often not unique and traditional wisdom says that explicit regularization such as weight decay must be added to ensure that the solution is stable or meaningful. However, such wisdom has been challenged by modern machine learning practice, where a small generalization error is achieved with massively \emph{overparameterized} ($p \gg n$) hypothesis classes $\mathcal{H}$ such as deep neural networks, without any explicit regularization. This implies that in such settings, the optimization method used for (\ref{eq:ls}) has a favorable \emph{implicit} bias towards a particular choice of $\vtheta_{opt} \in \mathcal{H}$ among all empirical risk minimizers. As neural networks can be trained with a particularly simple algorithm, (stochastic) gradient descent, a flurry of research in the past several years, starting with \cite{simonyan2014very, he2015delving, zhang2016understanding, canziani2016analysis}, has been devoted to answering the question: 
\begin{center}
    \emph{How and when does the implicit bias of gradient descent interact favorably with the structure} 
\end{center}
\begin{center}
    \emph{of a particular problem to achieve better performance in the interpolation regime? }
\end{center}

\bigskip
The papers \cite{belkin2018does} and \cite{liang2018just} observed that the power of overparameterization is not limited to neural networks, and can even be found in \emph{linear} interpolation models, where the feature basis $\{\psi_k\}_{k=1}^p$ is fixed, and the empirical risk is a quadratic function of the parameters: $ \| \Psi \vtheta  - {\bf y} \|^2 = \sum_{j=1}^n ( \sum_{k=1}^p \theta_k \psi_k({\bf x}_j) - y_j)^2$. In this setting, the implicit bias of gradient descent is well understood: by applying (stochastic) gradient descent to the empirical loss (\ref{eq:ls}) with initialization belonging to the range of the feature matrix $\Psi$, the solution converges to the parameter solution $\vtheta_{\text{min}}$ of minimal $\ell_2$-norm among all interpolating  solutions.\footnote{Observe that the gradient descent iterates $\vtheta_t$ remain in the row span of the feature matrix 
$\Psi$ if $\vtheta_0$ is in the row span, and the minimal-norm solution is the unique solution in the row span of the feature matrix.} 

The work \cite{belkin2019reconciling} claimed that improvement in generalization error is due to the connection between small $\ell_2$-norm of a parameter solution $\vtheta_{opt}$ and smoothness of the corresponding interpolating function $f_{\vtheta_{opt}}$. This connection was highlighted through the example of linear interpolation with random Fourier features \cite{RR08} where the feature basis functions 
$\psi_k: \mathbb{R}^d \rightarrow \mathbb{C}$ are random complex exponentials $\psi_k({\bf x}) = e^{i \langle{ \vw_k, {\bf x} \rangle}}$ with $\vw_k \sim \gN({\bf 0},\mI_d)$, and which can be viewed as a class of two-layer neural networks with fixed weights in the first layer. 
As the number of features $p \rightarrow \infty$, this basis converges to that of the reproducing kernel Hilbert space (RKHS) of smooth functions corresponding to the Gaussian kernel, and the interpolating solution by gradient descent converges to the smooth function with minimal RKHS norm. 

While this connection between the minimal $\ell_2$-norm and smoothness of the solution is intuitive and intriguing, a sharp analysis of the connection has not been provided. In this paper, we initiate this analysis by deriving exact non-asymptotic expressions for the generalization error and corresponding confidence bounds in a random Fourier series model which serves as a model for random smooth functions. We show precisely how a bias towards smooth interpolants, in the form of weighted trigonometric interpolation, results in a smaller generalization error in the overparameterized regime compared to the underparameterized regime. We note that in the linear setting and with equidistant sampling points, randomness in the model is required (here in the form of random Fourier coefficients), as  pathological counter-examples can be otherwise constructed.
Our analysis provides insight into the power of overparameterization in modern machine learning. 

\subsection{Main Contribution and Outline}
\begin{figure}[ht]
    \centering
    \includegraphics[width=.6\linewidth]{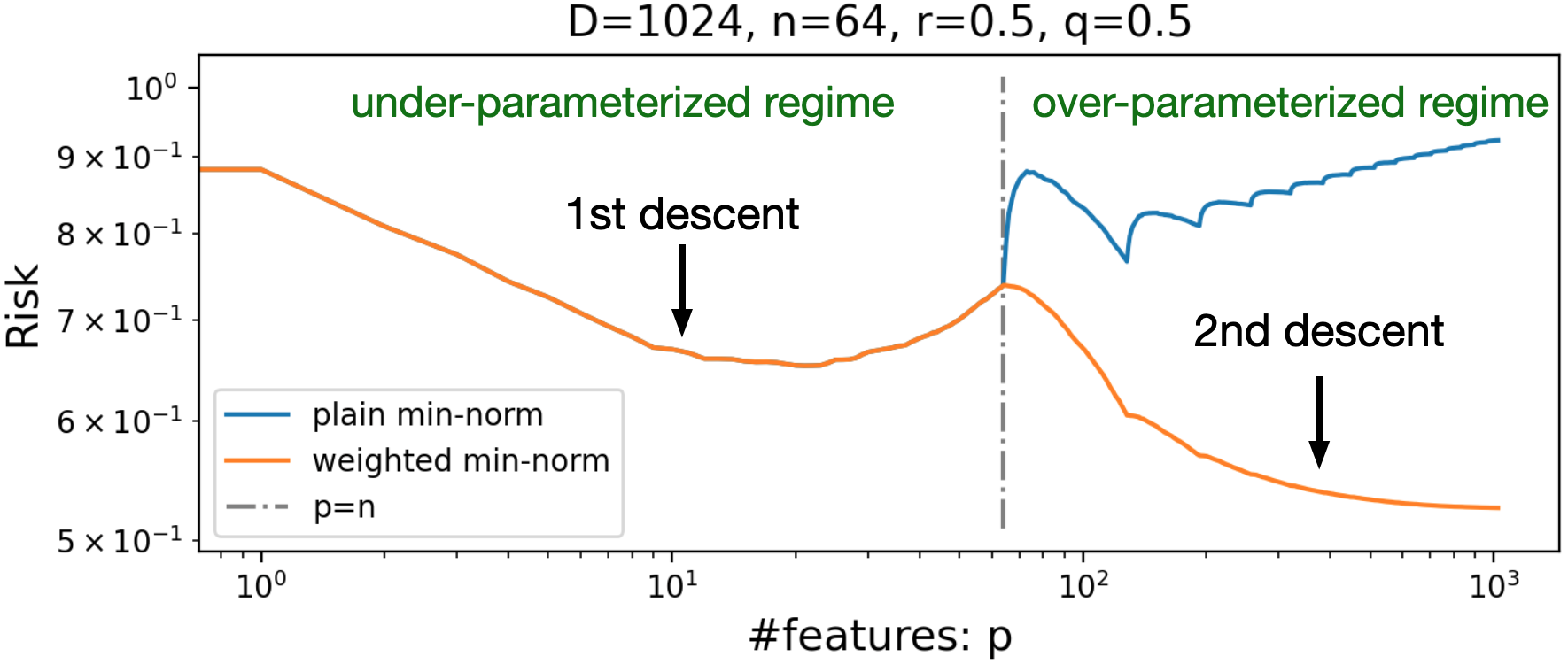}
    \vspace{-0.2cm}
    \caption{Demonstration of empirical risks ($\|\vtheta - \hat{\vtheta}\|^2$, average of 100 runs) comparing plain and weighted min-norm (with $q=0.5$) estimators. Here, $\vtheta$ is sampled from the distribution in Def. \ref{asump:x} for $D=1024, r=0.5, p \in [D]$.}  
    \label{fig:fourier_intro}
\end{figure}

In this paper, we consider the method of weighted $\ell_2$-norm trigonometric interpolation in Fourier feature space, where the weight on a particular feature is proportional to the $q$th power of its gradient norm to encourage lower-frequency features.  This additional degree of freedom, which is possible only in the overparameterized regime, enables us to reduce the risk.

We consider the weighted $\ell_2$-norm interpolation in Fourier feature space of equispaced training data $(x_j,f_{\vtheta}(x_j))$ corresponding to samples of a function with sharply decaying Fourier series $\vtheta$. Our key theoretical results are as follows. In Theorems \ref{thm:diff-equi} and \ref{thm:under}, we derive analytic expressions for the risk $\mathbb{E}[\|\vtheta-\widehat{\vtheta}\|_2^2]$ both in overparameterized and underparameterized regimes. The expression in the  overparameterized regime is particularly representative since $\vtheta-\widehat{\vtheta}$ concentrates well around its expectation, as shown in Theorem \ref{thm:highprob}. We then derive informative upper bounds for the risk in Theorem \ref{thm:order}. Moreover, Theorem \ref{thm:lower bound} states that with sufficient decay, the solution in the overparameterized regime is strictly better than that in the underparameterized regime. We illustrate the trends of empirical risk in Figure \ref{fig:fourier_intro} and detailed numerical results with different settings can be found in Figure \ref{fig:fourier}.

In Section~\ref{sec:formulation} we introduce the notation and precise formulation of the problem. In Section~\ref{sec:overpara} and \ref{sec:underpara} we analyze the risks in both over- and under-parameterized regimes, respectively. We provide extensive numerical results in Section~\ref{sec:exp} and discuss the importance of randomness in Section~\ref{sec:diss}. 
\subsection{Previous work on generalization and overparameterization}\label{sec:previous}
The work of \cite{belkin2019reconciling} initiated the study of the extended bias-variance trade-off curve and showed that this curve often exhibits double descent behavior, where the risk in the overparameterized regime $p \gg n$ can decrease to a point below the best possible risk in the underparameterized regime. They illustrated that this behavior naturally occurs in kernel regression/interpolation problems.

Subsequently, several works derived quantitative bounds on the risk in the interpolation setting but required that 
(a) the features are random (so that random matrix theory can be leveraged) or (b) $p$ and $n$ are in the asymptotic regime and go to infinity at a comparable rate (or both).  In contrast, our results are for deterministic Fourier features and hold for any $p$ and $n$. The precise high-dimensional asymptotic risk for a general random model with correlated covariates was derived in \cite{hastie2019surprises}. The work \cite{bartlett2020benign} derived sharp bounds on the risk in general linear regression problems with non-isotropic subgaussian covariates and highlighted the importance of selecting features according to higher-variance covariates. Other works include \cite{ghorbani2019linearized, mei2019generalization,geiger2020scaling, nakkiran2019deep}.

Prior to the above line of work, the seminal paper of Belkin, Hsu, and Xu  \cite{belkin2020two} introduced a discrete Fourier series model which is particularly amenable to empirical risk analysis, and which served as a main inspiration for this work.  The paper \cite{belkin2020two} provides a  theoretical analysis for the discrete Fourier series model in the setting of randomly chosen Fourier features, unweighted optimization, and isotropic covariates, in the asymptotic setting. Empirical evidence pointing to improved generalization using weighted optimization and Fourier series truncated to the lowest frequencies instead of random  frequencies was also provided in \cite{belkin2020two}, but without theoretical analysis. In this sense, our paper can be viewed as answering an open question regarding the role of weighted optimization in Fourier series interpolation from \cite{belkin2020two}. Throughout this paper, we refer to the solution by unweighted optimization as considered in  \cite{belkin2020two} as ``plain min-norm estimator" and provide a detailed comparison between their estimator and our weighted min-norm estimator.

We acknowledge a concurrent preprint \cite{li2020generalization} that also studies generalization error of minimum weighted norm interpolation, but focusing on upper bounds in the asymptotic setting from an approximation theory viewpoint.  By contrast, our work provides exact non-asymptotic expressions 
and bounds in probability 
for the generalization error.

\section{Formulation}\label{sec:formulation}
Smooth functions are characterized by the rate of decay in their Fourier series coefficients---the smoother the function, the faster the decay.\footnote{The classical Sobolev spaces are Hilbert spaces defined in terms of Fourier series whose coefficients decay sufficiently rapidly. For square-integrable complex-valued functions $f$ on the circle $\mathbb{T}$, consider the space of functions
$ H^r(\mathbb{T}) = \{ f \in L^2(\mathbb{T}): \| f \|_{r,2}^2 := \sum_{j = -\infty}^{\infty} (1 + |j|^2)^{r} |\hat{f}(j)|^2 < \infty\}, r \in \sR, r\geq 0,$  where $\hat{f}(j)$ is the $j$th Fourier series coefficient of $f$. If $r \in \sN$, by duality between differentiation in time and multiplication in frequency,  the Sobolev norm is equivalently defined in terms of the $r$th derivative $f^{(r)}$: $\| f \|_{r,2}^2 = \| f \|_{L_2}^2 + \| f^{(r)} \|_{L_2}^2.$}
Drawing inspiration from this connection, we consider as a model for random smooth periodic functions the class of trigonometric polynomials $f: [-\pi,\pi] \rightarrow \mathbb{C}$ with random $r$-decaying Fourier series coefficients: 

\begin{definition} \label{asump:x} 
\textbf{(Random Fourier series with $r$-decaying coefficients)}
Fix $D \in \mathbb{N}$ and $r \geq 0$. We say that a function is a random Fourier series with $r$-decaying coefficients if\footnote{
For ease of exposition, we only work with positive frequencies, although it may seem more natural to work with trigonometric polynomials of the form $f(x) = \sum_{k=-D}^D \theta_k e^{ikx}$. All our results can be formulated within that setting, by symmetrically extending the weights to negative indices $k$ and by replacing $D$ with $2D-1$, and similarly for $n$ and $p$. The notation, however, will be more heavy and then make the comparison with other works less straightforward.} 
\begin{align}
f_{\vtheta}(x) = \sum_{k=0}^{D-1} \theta_k e^{ i k x}, \quad
0 \leq x \leq 2 \pi,
\end{align} 
where $\vtheta \in \mathbb{C}^D$ is a random vector satisfying $\ex [\vtheta] = {\bf 0}$ and $\ex[\vtheta \vtheta^{*}] = c_r \Sigma^{2r}$, where
$\Sigma := \text{diag}\left((k+1)^{-1}, k = 0,\dots, D-1\right) \in \sR^{D \times D}$ and $c_r = (\sum_{k=0}^{D-1} (k+1)^{-2r})^{-1} \st \ex [\| \vtheta \|^2] = 1$.
\end{definition}

We observe $n \leq D$ training samples $(x_j, y_j)_{j=0}^{n-1} = (x_j, f_{\vtheta}(x_j))_{j=1}^n$ of such a function $f_{\vtheta}$ at equispaced points on the domain, $x_j = \frac{2 \pi j}{n}$, $j \in [n]$, where $[n]$ denotes the set $\{0,\dots,n-1\}$ for notation simplicity. We can express the observation vector ${\bf y} \in \mathbb{C}^n$ concisely as ${\bf y} = \mF \vtheta $ in terms of the \emph{sample discrete Fourier matrix} $\mF \in \mathbb{C}^{n \times D}$ whose entries are $(\mF)_{j,k} = e^{i k x_j } = e^{2 \pi i j k /n}$, $j\in [n]$, $k\in [D]$. If $D$ is a multiple of $n$, i.e., $D= \tau n$ for $\tau\in \sN$, then we can write $\mF = [\mF^{(n)}|\mF^{(n)}| \cdots | \mF^{(n)}]$, where $\mF^{(n)} \in \sC^{n \times n}$ is the discrete Fourier matrix in dimension $n$.

We fit the training samples to a degree-$p$ trigonometric polynomial $\hat{f}_{\hat{\vtheta}}(x) = \sum_{k=0}^{p-1} \widehat{\theta  }_k e^{ i k x}$ such that $\hat{f}_{\hat{\vtheta}}(x_j) \approx f_{\vtheta}(x_j)$, $j \in [n]$, i.e., ${\bf y} \approx \mF_T \widehat{\vtheta}_T$ and $\vtheta_{T^c} = \v0$, where $\mF_T \in \mathbb{C}^{n \times p}$ is the matrix containing the first $p$ columns of $\mF$, indexed by $T=[p]$. We solve for $\widehat{\vtheta}_T$ as the least squares fitting vector in the regression regime $p < n,$ and as the  solution of minimal weighted $\ell_2$ norm in the interpolation regime $p > n$:  
\begin{align}
\label{eq:opt}
\widehat{\vtheta}_T = 
\left\{ \begin{array}{ll} 
\text{arg} \min_{\vw \in \mathbb{C}^p} \| \mF_{T} \vw - \vy \|_2^2;  &  p \leq n \\
\text{arg} \min_{\vw \in \mathbb{C}^p} \| \Sigma_T^{-q} \vw \|_2^2 \quad \st \mF_T \vw = {\bf y} & p > n \\
\end{array} \right\},
\end{align}
where $\Sigma_T \in \mathbb{R}^{p \times p}$ is the diagonal matrix as in Def.~\ref{asump:x} restricted to its first $p$ rows and $p$ columns and $q\geq 0$ controls the rate of growth of the weight. Note that the weight matrix $\Sigma_T^{-q}$ has no influence on the estimator in the underparameterized regime $p\leq n$. Denoting by $\mA^\dagger$ the Moore-Penrose pseudo-inverse of a matrix $\mA$, we can write the solution in both the under- and overparameterized case as $\widehat{\vtheta}_T = \Sigma_T^{q} (\mF_T \Sigma_T^q)^\dagger {\bf y}.$

We will derive sharp non-asymptotic expressions for the risk of the estimator $f_{\htheta}$ in terms of $p, n, D, r,$ and $q$. 
The risk in this setting is 
defined as
\begin{align}
\label{eq:risk}
\risk_q = \mathbb{E}_{\vtheta } \left[ \int_{-\pi}^{\pi} | f_{\vtheta}(x) - f_{\htheta}(x) |^2 dx \right] = \ex_{\vtheta } \left[\| \vtheta  - \htheta \|_2^2\right],
\end{align}
where the last equality follows from Parseval's identity.

\section{Risk Analysis for Decaying Fourier Series Model in the Overparameterized Setting}\label{sec:overpara}
We first derive non-asymptotic expressions, upper bounds, and concentration for the risk \eqref{eq:risk} via  plain \cite{belkin2020two} $(q = 0)$ and weighted $(q > 0)$ $\ell_2$ regression with Fourier series features in the overparameterized regime ($p \geq n$). For ease of exposition, we restrict to the case where $p$ is an integer multiple of $n$, but note that the result can be extended to the general case. Finally, we introduce the notation 
\begin{align}
\label{eq:notation} 
t_j &:= (j+1)^{-1}, \quad j \in [D], \nonumber \\
\Sigma &:= \diag([t_0,\dots,t_{D-1}]), \nonumber \\
c_r &:= 1/\sum_{j=0}^{D-1} t_j^{2r}.
\end{align}
\begin{theorem} \label{thm:diff-equi} 
\textbf{(Risk in overparameterized regime)}
Assume $D = \tau n$ and $p = l n$ for $\tau \geq l$, $\forall \tau, l \in \sN_+ :=\{1,2,\dots\}$ ($p\geq n$). Let the feature vector $\vtheta$ be drawn from a distribution with $\ex[\vtheta] = \v0$ and $\ex[\vtheta \vtheta^*]  = c_r\Sigma^{2r}$. Then the risk \eqref{eq:risk} for the regression coefficients $\hat{\vtheta}$ fitted by plain min-norm estimator ($q =0$) and weighted min-norm estimator $(q > 0)$  are 
\begin{align}
\risk_0 &= 1 - \frac{n}{p} + \frac{2n}{p} \cdot \crs \sum_{j=p}^{D-1} \tjr; \\ 
\risk_q & = 1  - { \color{black}  2} \crs \sum_{k=0}^{n-1} \frac{ \sinsuma{ {2q+2r} } }{ \sinsuma{ {2q} } }   + 
 \crs \sum_{k=0}^{n-1} \frac{ \p{ \sinsuma{ {4q} } } \p{ {\color{black} \sum_{\nu=0}^{\tau-1} } t_{k+n\nu}^{2r} }  }{ \p{ \sinsuma{ {2q} } } ^2 } 
\label{eq:risks}
\end{align}
\end{theorem}

\begin{remark}\label{rmk:risk} While the general risk expressions are difficult to parse, special cases are straightforward: if $p=D (l=\tau)$, then $\risk_0 = 1 - n/D$. And if $n=p=D$ ($l = \tau = 1$), {\color{black}then $\risk_q = 1 - \crs \sum_{k=0}^{D-1} t_k^{2r} = 0, \forall q$}, so that $\risk_0 = \risk_q = 0$.  We provide a more informative asymptotic rate in Theorem \ref{thm:order} to show the decrease of risk as $p$ and $n$ increases.
\end{remark}

\begin{remark} \label{rmk:infinited}
{\color{black} When $r>1/2$, our results also hold in the limit $D \rightarrow \infty$ corresponding to a Fourier series expansion rather than a finite trigonometric sum, as the series $\sum_j t_j^{2r}$ converges. In case $r<1/2$, we impose finite $D$ as we run into convergence issues with infinite series. However, in the unweighted case $r=0$, where we have an explicit expression for the infinite series involved, we may also take $D \rightarrow \infty$ as expressions cancel appropriately.  Thus, the restriction to finite $D$ in the range $ 0 < r \leq 1/2$ is likely unnecessary.  Nevertheless, we don't have explicit expressions for the diverging sums as we have in the case $r=0$, and so removing the finite $D$ restriction in this range is challenging.}
\end{remark}

Using the expressions in Theorem \ref{thm:diff-equi}, we can quantify how smoothness (as reflected in the rate of decay $r>0$ in the underlying Fourier series coefficients) can be exploited by setting the weights accordingly to reduce the risk in the overparameterized setting. 

\begin{proof}[Proof of Theorem \ref{thm:diff-equi}] \label{pf:diff-equi}

The proof is based on the following lemmas {\color{black}. See Appendix \ref{app:lemmas} for proof of all the lemmas}. Below, the matrix $\mF_{T^c} \in \sC^{n \times (D-p)}$ is the submatrix of $\mF$ with the columns in $T^c = [D] \setminus T$.

\begin{lemma} \label{lem:weighted} \textbf{(Risks of estimators in overparameterized regime)} Assume $p \geq n$ and let the feature vector $\vtheta \in \sC^D$ be drawn from a distribution with $\ex[\vtheta] = \v0$ and $\ex[\vtheta \vtheta^*] = \mK$, where $\mK$ is a diagonal matrix. The risk of the weighted min-norm estimator with $q \geq 0$ is  {\color{black}$\risk_q = \ex [ \|\vtheta - \hat{\vtheta} \|^2] = \tr{\mK } - 2 \gP_q+ \gQ_{q,1} + \gQ_{q,2}$}, where
$ \gP_q = \tr{ \Fst  \sTqs \kt \Fst^* (\Fst \sTqs \Fst^*)^{-1} } $, {\color{black}$\gQ_{q,1} =  \text{tr} (\Fst \sT^{4q} \Fst^*(\Fst \sTqs \Fst^*)^{-1} $ $\Fst \kt \Fst^*$  $(\Fst \sTqs \Fst^*)^{-1} )$}, and
${\color{black}\gQ_{q,2} } = \text{tr}( \Fst \sT^{4q} \Fst^*$ $(\Fst \sTqs \Fst^*)^{-1}\Fstc \ktc \Fstc^* (\Fst \sTqs \Fst^*)^{-1})$.
\end{lemma}

\begin{lemma}\label{lem:equi-fst} \textbf{(Properties of $\Fst$)} 
Assume that $D=\tau n$ and $p = n l$ for $\tau, l \in \sN_+$. Then, $\Fst \Fst^* = p \mI_n$. For $u \in \sN_+$, define $\mA_u :=  \Fst \sT^u  \Fst^*$ and $\mC_u:= \Fstc \sTc^u \Fstc^*$, where $\Sigma$ is a diagonal matrix (e.g., $\Sigma$ in Def. \ref{asump:x}). Then, $\mA_u$ and $\mC_u$ are circulant matrices.
\end{lemma}
Since $\mau$ and $\mcu$ are circulant matrices, $\mau = \un \Lambda_{a,u} \un^*$ and $\mcu = \un \Lambda_{c, u} \un^*$, where $\un$ is the unitary discrete Fourier matrix of size $n$, and $\Lambda_{a,u}$ and $\Lambda_{c,u}$ are diagonal matrices with eigenvalues of $\mau$ and $\mcu$ on the diagonal, respectively. The eigenvalues can be calculated by taking discrete Fourier transform of the first column of $\mau$ or $\mcu$. 

Let $\fn$ be the $n$th order discrete Fourier matrix, i.e., $(\fn)_{s,j}= \omn^{sj}$, then for any $s\in [n]$, the $s$th diagonal element (eigenvalue) of $\Lambda_{a,u}$ or $\Lambda_{c,u}$ is 
\begin{align*}
\begin{split}
    \lambda_{a,u}^{(s)} & = \fn[s,:]\mau[:,0] =  \sum_{j=0}^{n-1} \omn^{s j} \p{ \aj } =   \las{u} \\
    \lambda_{c,u}^{(s)} & = \fn[s,:]\mcu[:,0] =  \sum_{j=0}^{n-1} \omn^{s j} \p{ \cj } =  \lcs{u}
\end{split}
\end{align*}
For $s,k \in [n]$ we define
\begin{align}
    \esk := \sum_{j=0}^{n-1}  \omn^{(s-k) j} =
    \begin{cases}
      n, & \text{ if } k=s, \\
 0 , & \text{otherwise}. 
  \end{cases}
\end{align}
If the random Fourier series has $r$-decaying coefficients, i.e., $\mK = \covr $ ($r \geq 0$), by Lemma~\ref{lem:weighted},
\begin{align*}
\begin{split}
     \gP_q & = \crs  \tr{ \Fst  \sT^{2q+2r } \Fst^* (\Fst \sTqs \Fst^*)^{-1} }  = \crs  \tr{ \un \Lambda_{a,2q+2r } \un^*\un \Lambda_{a,2q}^{-1} \un^*} \\
    & =  \crs  \tr{\Lambda_{ {a,2q+2r } } \Lambda_{a, 2q}^{-1}} 
    = \crs  \sum_{s=0}^{n-1} \frac{ \suma{ {2q+2r} } \esk }{ \suma{ {2q} } \esk }  = \frac{1}{ \sumr }\sum_{k=0}^{n-1} \frac{ \sinsuma{ {2q+2r} } }{ \sinsuma{ {2q} } },\\  
    \gQ_{q,2} & = \crs  \tr{\Fstc \sTcr \Fstc^* (\Fst \sTqs \Fst^*)^{-1} \Fst \sT^{4q} \Fst^*  (\Fst \sTqs \Fst^*)^{-1} } \\
    & = \crs  \tr{\un \Lambda_{c,2r } \un^* \un \Lambda_{a,2q}^{-1} \un^* \un \Lambda_{a,4q} \un^* \un \Lambda_{a,2q}^{-1} \un^* } 
    = \crs  \tr{ \Lambda_{c,2r } \Lambda_{a,2q}^{-1} \Lambda_{a,4q} \Lambda_{a,2q}^{-1} } \\
    & = \crs  \sum_{s=0}^{n-1} \frac{ \p{ \suma{ { 4q } } \esk } \p{ \sumc{ { 2r } } \esk }   }{ \p{ \suma{ { 2q } } \esk } ^2  } \\
    & = \crs \sum_{k=0}^{n-1} \frac{ \p{ \sinsuma{ {4q} } } \p{ \sinsumc{ {2r} } }  }{ \p{ \sinsuma{ {2q} } } ^2 }.
\end{split}
\end{align*}
{\color{black}Similarly, we have $\gQ_{q,1} = \crs \sum_{k=0}^{n-1} \frac{ \p{ \sinsuma{ {4q} } } \p{ \sinsuma{ {2r} } }  }{ \p{ \sinsuma{ {2q} } } ^2 }$},
and the risk satisfies
\begin{align*}
\begin{split}
     \risk_q & =  { \color{black} 1- 2\gP_q + \gQ_{q,1} + \gQ_{q,2} } \\
 & = 1  - {\color{black}2 } \crs \sum_{k=0}^{n-1} \frac{ \sinsuma{ {2q+2r} } }{ \sinsuma{ {2q} } }   + 
 \crs \sum_{k=0}^{n-1} \frac{ \p{ \sinsuma{ {4q} } } \p{ {\color{black}\sum_{\nu=0}^{\tau-1}} t_{k+n\nu}^{2r} }  }{ \p{ \sinsuma{ {2q} } } ^2 }  .
\end{split}
\end{align*}
Let $q=0$, 
\begin{align*}
    \risk_0
 = 1- 2\crs \sum_{k=0}^{n-1} \frac{ \sum_{\nu=0}^{l-1} t_{k+n\nu}^{2r}  }{ l } + c_r \sum_{k=0}^{n-1} \frac{ \sum_{\nu=0}^{\tau-1} t_{k+n\nu}^{2r}  }{ l }  = 1- \frac{n}{p} + \frac{2n}{p} \cdot \frac{ \sum_{j=p}^{D-1} \tjr  }{ \sumr } .
\end{align*}

\end{proof}

{\color{black}
\begin{remark}
\label{rmk:highd-over} 
Assume equispaced points $\vx \in \sR^d$, and let $\mF_T$ be the matrix with first $p$ features for each dimension. Let $\mH_{T,d} = \Fst^{\otimes d} := \Fst \otimes \cdots \otimes \Fst  \in \sC^{n^d \times p^d}$, $\vtheta \in \sC^{p^d}$, $\vy \in \sC^{n^d}$, and $\Sigma_{[\vk]} \propto \|\vk\|_2$ with where $\vk := [k_1, k_2, \dots, k_d]^T$. Using the same reparameterization $\vbeta = \Sigma_{T,d}^{-q} \vtheta$, we have
$\htheta_T = \Sigma_{T,d}^{q} (\mH_{T,d} \Sigma_{T,d}^{q})^{\dagger} \vy$. Similarly, the matrices  $\mA_u := \mH_{T,d} \Sigma_{T, d}^{u} \mH_{T,d}^* $ and $\mC_u := \mH_{T^c,d} \Sigma_{T^c, d}^{u} \mH_{T^c,d}^* $ are diagonalizable since they are block circulant matrices with circulant blocks ($\mF_T$ or $\mF_{T^c}$ ) \cite{combescure2009block}. Hence, the risk for the weighted min-norm estimator in the overparameterized case is straightforwardly generalized to the high-dimensional setting. Figure \ref{fig:interp2d} shows an example comparing plain and weighted min-norm interpolation in a $2$-dimensional example.
\end{remark}
}
\subsection{Asymptotic Rate of Weighted Min-norm Risk}
In this section, we will derive an informative upper bound for \eqref{eq:risks} in Theorem \ref{thm:order} that demonstrates the asymptotic behavior of the risk and its relation to the parameters $n$, $p$, and $r$. {\color{black} Proof of Theorem \ref{thm:order} and related lemmas can be found in Appendix \ref{app:theorems}.}
\begin{theorem}\label{thm:order} \textbf{(Asymptotic rate of weighted min-norm risk)}
In the overparameterized setting of Theorem \ref{thm:diff-equi}, if $q=r>1/2$ and $p = nl$ with $l \in \sN_+, l \geq 2$, then the risk of weighted optimization satisfies 
\begin{align}\label{eq:rate}
    \risk_q \leq a n^{-2r+1} + b n^{-2r} p^{-2r+1}
\end{align}
with $a = \frac{2+d_r n^{-2r}}{(1+d_rn^{-2r})(1-(D+1)^{-2r+1})}$, $b=\frac{d_r}{(1+d_r n^{-2r})(1-(D+1)^{-2r+1})}$, and
$d_r=\frac{2^{-2r+1}-(l+1)^{-2r+1}}{2r-1}$.

\end{theorem} 
\begin{remark}\label{rmk:rate} {\color{black} As $\#$features $p$ and $\#$samples $n$ increase, the risk of weighted min-norm estimator asymptotically converges to zero at the above rate.} Indeed, for sufficiently large $D$ and $n$, the constants in the above theorem satisfy $a\leq 2$, and $b\leq2^{-2r+1}/{(2r-1)}$, so thus 
\begin{align}\label{eq:larged}
    \risk_q \leq 2n^{-2r+1} + \frac{2}{2r-1}(2n)^{-2r}p^{-2r+1}.
\end{align}
\end{remark}
\subsection{Concentration of Error}
In this section, we show that the risk does not deviate far from its expectation with high probability, due to concentration of the error $\vtheta-\widehat{\vtheta}$.  Proof of {\color{black}Theorem \ref{thm:highprob} is in Appendix \ref{app:theorems}.}

\begin{theorem} \label{thm:highprob} \textbf{(Probabilistic Bound)} 
In the overparameterized setting of Theorem \ref{thm:diff-equi}, suppose $r \geq q\geq \frac{1}{2}$ and suppose $\vtheta$ has independent sub-Gaussian coordinates with $\|\vtheta_k\|_{\psi_2} = \sqrt{c_r}k^{-r}$, where $\|X\|_{\psi_2}:=\inf\{t>0:\mathbb{E}(\text{exp}(X^2/t^2))>2\}$ is the sub-Gaussian norm. For any $t > 0$, 
\begin{align}
  \sP \p{ \left|\|\vtheta  - \hat{\vtheta}\|_2^2  - \mathbb{E}\|\vtheta  - \hat{\vtheta}\|_2^2 \right|  \leq t } \geq 1-2\text{exp}\left[-\min\left(\frac{t^2}{T_q^2}, \frac{t}{T_q}\right)\right], 
\end{align}
where $T_q= 4(2r-1)\sqrt{\frac{q(24q^2 - 17q + 3)}{(2q-1)^2(4q-1)}}$.
\end{theorem}

\section{Risk Analysis in the Underparameterized Setting and Benefits of Overparameterization}\label{sec:underpara}
In order to fully understand the benefit of overparameterization, we derive the non-asymptotic risk for the estimators in the underparameterized regime $(p\leq n)$, where $q$ does not have an influence on the estimators and, hence, on the risk.

\begin{theorem} \label{thm:under} \textbf{(Risk in underparameterized regime)} 
Suppose $D = \tau n$ for $\tau \in \sN_+$. Suppose $p\leq n$, and assume that the feature vector $\vtheta$ is drawn from a distribution with $\ex[\vtheta] = \v0$ and $\ex[\vtheta \vtheta^*]  = c_r\Sigma^{2r}$.
Then the risk ($\ex  [ \| \vtheta - \hat{\vtheta}\|^2] $) is given by 
 \begin{align}
    \begin{split}
    \risk_{under}  = \crs \p{ \sum_{j=p}^{D-1} \tjr + \sum_{k=1}^{\tau-1} \sum_{j=0}^{p-1} t_{kn+j}^{2r} } .
    \end{split}
\end{align}
\end{theorem}

\begin{remark}\label{rmk:trend_under}
When $r=0$, $\risk_{under} = \frac{1}{D}(D-p + (\tau-1)p)= 1+p(\frac{1}{n} - \frac{2}{D})$ and the risk increases with $p$ until $p=n$, provided $n <D/2$. From Figure \ref{fig:fourier}, as we vary $r$ in the range $0 \leq r \leq 1$, this behavior persists for a while, then changes to a $U$-shape curve, and lastly to a decreasing curve. For $r \geq 1$, we prove that the risk is monotonically decreasing in $p$. 
\end{remark}

\begin{proof} [Proof of Theorem~\ref{thm:under}]The proof of Theorem~\ref{thm:under} is based on Lemma \ref{lem:ls}
\begin{lemma} \label{lem:ls} \textbf{(Risks of weighted and plain min-norm estimator in the underparameterized regime)} 
Let the feature vector $\vtheta$ be sampled from a distribution with $\ex[\vtheta] = \v0$ and $\ex[\vtheta \vtheta^*] = \mK$. In the underparameterized regime ($p \leq n$), the regression coefficients $\hat{\vtheta}$ are fitted by weighted least squares with $\Sigma^q$ as the re-parameterization matrix, then for any $q \geq 0$, $\hat{\vtheta}_T =  (\Fst^* \Fst)^{-1} \Fst^* \vy, \hat{\vtheta}_{T^c} = \v0$, where $\vy = \Fst \vtheta_T + \Fstc \vtheta_{T^c}$. The risk is given by
\begin{align*}
    \risk_{under} = \ex [\| \vtheta - \hat{\vtheta} \|^2 ] = \tr{ \mK_{T_c} } + \tr{  \Fst (\Fst^* \Fst)^{-2} \Fst^* \Fstc \mK_{T_c} \Fstc^* } . 
\end{align*}
\end{lemma}
Denote $\omn = \exp(-\frac{2\pi i}{n})$, for any $k_1,k_2 \in [p]$ with $p < n$,
\begin{align*}
  (\Fst^* \Fst)_{k_1, k_2} & = \sum_{j=0}^{n-1} \exp  \p{ - \frac{2\pi i}{n} k_1\cdot j} \exp \p{ \frac{2\pi i}{n} k_2\cdot j}   =  \sum_{j=0}^{n-1} \omn^{(k_1-k_2)\cdot j} 
  = \begin{cases}
     n, & \text{if } k_1 = k_2, \\
    0  , & \text{otherwise}. 
  \end{cases}
\end{align*}
Moreover, for $0 \leq k_1, k_2 \leq D-p-1$, we have %
\begin{align*}
  (\Fstc^*  \Fstc)_{k_1, k_2} & = \sum_{j=0}^{n-1} \exp  \p{ - \frac{2\pi i}{n} (k_1+p)\cdot j} \exp \p{ \frac{2\pi i}{n} (k_2+p)\cdot j}   =  \sum_{j=0}^{n-1} \omn^{(k_1-k_2)\cdot j} \\
 & = \begin{cases}
     n, & \text{if} \quad  \exists \gamma\in \sN, ~ \st ~ k_1 - k_2 = \gamma n, \\
    0  , & \text{otherwise}. 
  \end{cases}
\end{align*}
Since $D= \tau  n$ and $0 < p<n$ it holds $v = D-p - n \cdot  \lfloor \frac{D-p}{n} \rfloor = n-p $ and $p= n-v$. Introducing the matrices $\mM_{n \times v} =  \begin{bmatrix}  \mI_{v \times v} \\ \mO_{p \times v }  \end{bmatrix}$, 
$\mN_{v \times n} =  \begin{bmatrix}  \mI_{v \times v} & \mO_{v \times p }  \end{bmatrix}$, 
$\mI_{n,v} = \mM_{n \times v} \mN_{v \times n} =  \begin{bmatrix}  \mI_{v \times v} & \mO_{v \times p }  \\ \mO_{p \times v} & \mO_{p \times p }  \end{bmatrix}$, we can write
\begin{align*}
    \begin{split}
        (\Fstc^* \Fstc)^2 & = n^2    \begin{bmatrix} 
        \mI_n   & \cdots & \mI_n  & \mM_{n \times v} \\
        \vdots  & \ddots & \vdots  & \vdots \\
        \mI_n  & \cdots & \mI_n  & \mM_{n \times v} \\
        \mN_{v \times n}  & \cdots & \mN_{v \times n}  & \mI_{v \times v} 
        \end{bmatrix} ^2\\
        & = n^2 
     \begin{bmatrix} 
       (\tau  -1) \mI_n + \mI_{n,v}   & \cdots &  (\tau  -1) \mI_n  + \mI_{n,v}   &   \tau  \mM_{n \times v} \\
        \vdots  & \ddots & \vdots  & \vdots \\
        (\tau  -1)   \mI_n + \mI_{n,v}   & \cdots &  (\tau  -1) \mI_n + \mI_{n,v}  & \tau  \mM_{n \times v}  \\
    \tau  \mN_{v \times n}  & \cdots &   \tau  \mN_{v \times n}   &    \tau  \mI_{v\times v}
        \end{bmatrix}  \triangleq  n^2 \mL.
    \end{split}
\end{align*}
Since $\Fst \Fst^* + \Fstc \Fstc^* = D \mI_n$ the risk is given as
\begin{align*}
    \ex  [\|\vtheta - \hat{\vtheta}\|^2]  & =  \crs  \tr{\sTcr} +  \frac{ \crs }{n^2}  \tr{ ( D \mI_n - \Fstc \Fstc^* )  \Fstc \sTcr \Fstc^*} \\ 
    & =  \crs  \tr{\sTcr} + \frac{ \crs }{n^2} \tr{ \p{ D \Fstc^* \Fstc - (\Fstc^* \Fstc)^2 } \sTcr }  \\
    & =  \crs  \p{1 + \tau  } \tr{\sTcr} -  \crs  \tr{\mL \sTcr} \\
    &=   \crs  \p{1 + \tau  } \tr{\sTcr} - \crs \tau \tr{ \sTcr } +   \crs  \sum_{k=1}^{\tau -1} \sum_{j=0}^{p-1} t_{kn+j}^{2r}   \\
    & =  \crs  \sum_{j=p}^{D-1} \tjr +  \crs  \sum_{k=1}^{\tau -1} \sum_{j=0}^{p-1} t_{kn+j}^{2r}  = \frac{ \sum_{j=p}^{D-1} \tjr + \sum_{k=1}^{\tau -1} \sum_{j=0}^{p-1} t_{kn+j}^{2r} }{ \sumr }.
\end{align*}

\end{proof}

Finally, we show that the ``second descent" of the weighted min-norm estimator in the overparameterized regime achieves a lower risk than in the underparameterized regime, provided $q\geq r \geq 1$.  In other words, it is where ``over- is better than under-parameterization". 

\begin{theorem} \label{thm:lower bound}
\textbf{(The Lowest risk)} In the setting of Theorems \ref{thm:diff-equi} and \ref{thm:under}, if $q \geq r \geq 1$, then 
\begin{itemize}
    \item[(a)] In the underparameterized regime ($p\leq n$), the risk is monotonically decreasing in $p$ and the lowest risk in this regime is $ risk_{under}^* = 2 \crs \sum_{j=n}^{D-1} t_j^{2r}$.
    \item[(b)] The lowest risk in the overparameterized regime ($p>n$) is strictly less than the lowest possible risk in the underparameterized regime. 
\end{itemize}
\end{theorem}

\begin{remark}\label{rmk:lowrisk}
While the above theorem holds for any $q$ satisfying $q \geq r \geq 1$, our experiments suggest that $q=r$ is an appropriate choice for any $r\geq 0$, corresponding to the case where the assumed smoothness $q$ employed in the weighted optimization matches the true underlying smoothness $r$. {\color{black} Another intuition for the good behavior around $q=r$ is $\gP_q - \gQ_{q,1}=0$. (See proof of Theorem \ref{thm:order} in Appendix \ref{app:theorems}.) For a range of choices for $r$ and $q$, the plots of the theoretical extended risk curves (fixed $n$, varying $p$) can be found in Appendix \ref{exp:other}. And the risk with $q \approx r$ in Figure \ref{fig:q_approx_r} in Appendix \ref{exp:other} shows the robustness of weighted optimization.}
\end{remark}

\section{Experiments}\label{sec:exp}
\paragraph{Discrete Fourier Models} In this experiment, we use Fourier series models $\mF \in \sC^{n\times D}, D=1024, n=64$ with $r$-decaying multivariate Gaussian  coefficients ($r=0.3, 0.5, 1.0$). 
$\Fst \in \sC^{n \times p}, {\color{black}\forall p \in [D]}$ is the observation matrix. The weighted min-norm estimator uses $\Sigma^q$, $q \geq 0$ to define the weighted $\ell_2$-norm. The theoretical curves are the risks calculated according to Theorems \ref{thm:diff-equi} and \ref{thm:under} on {\color{black} $p<n$ in the underparameterized regime and $p=l n, l=1,2,\dots, \tau$ in the overparameterized regime}. The empirical mean curves and the $80\%$ confidence intervals (CI) are estimated {\color{black}on all $p \in [D]$} by $100$ runs of independently sampled feature vectors $\vtheta$. 

\begin{figure}[ht]
    \centering
    \includegraphics[width=.98\linewidth]{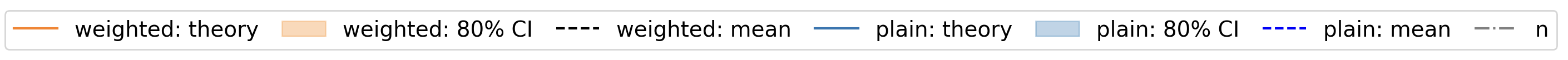}
    \includegraphics[width=.32\linewidth]{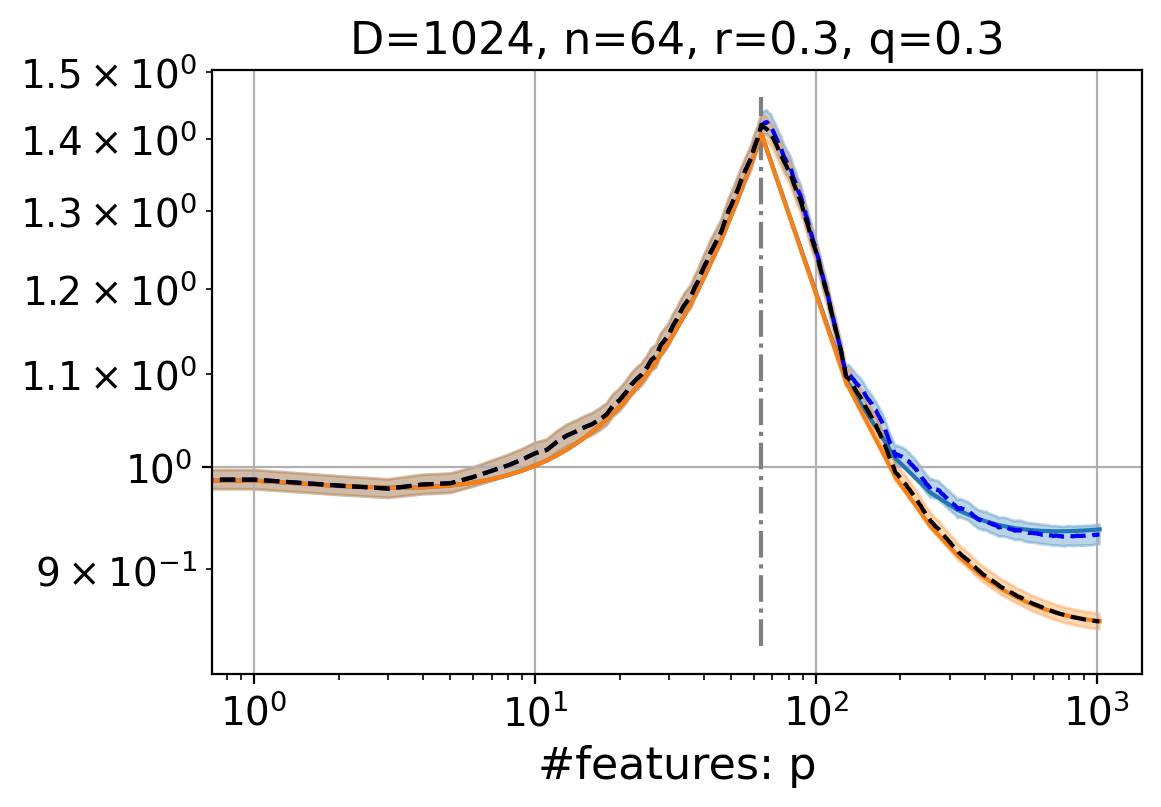}
    \includegraphics[width=.32\linewidth]{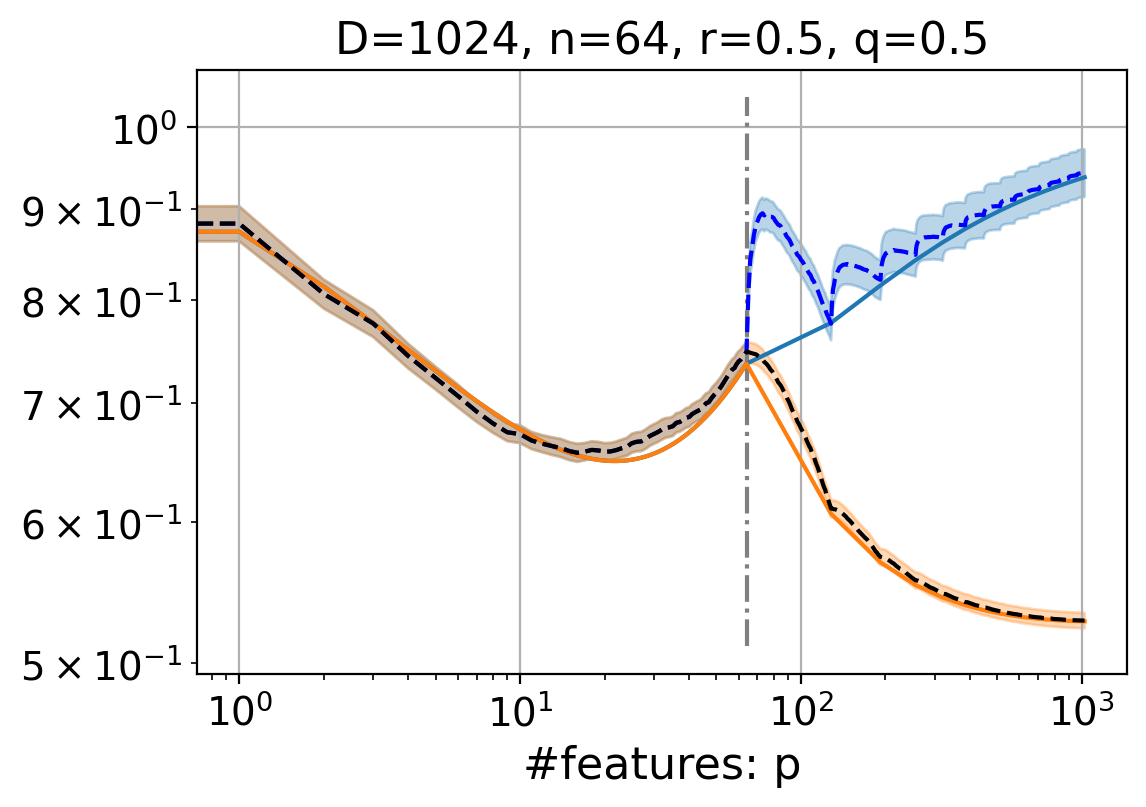}
    \includegraphics[width=.3\linewidth]{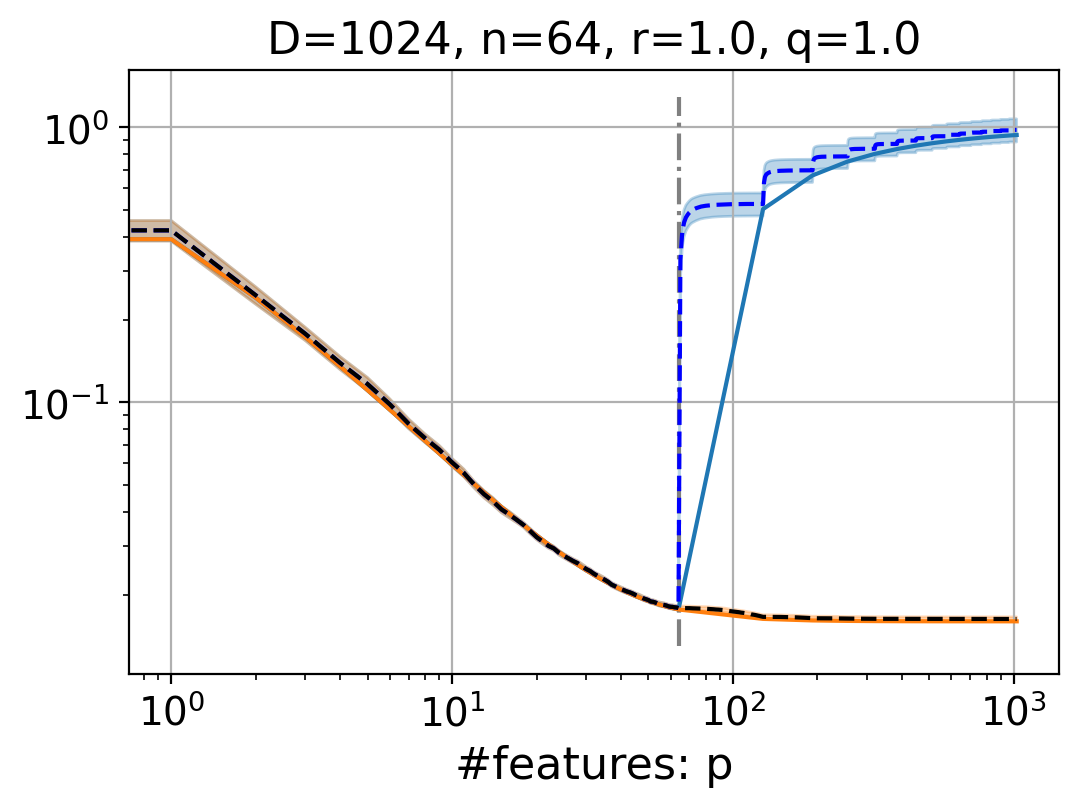}
    \vspace{-.4cm}
    \caption{Theoretical and empirical risks ($\|\vtheta - \hat{\vtheta}\|_2^2$) of plain and weighted min-norm estimators in log-log scale. Left to right: $r=q=0.3, 0.5, 1.0$.} 
    \vspace{-.4cm}
    \label{fig:fourier}
\end{figure}
Figure \ref{fig:fourier} shows that the empirical mean risks match the theoretical risks $\ex [\|\vtheta - \hat{\vtheta}\|^2]$ of Theorems \ref{thm:diff-equi} and \ref{thm:under} very accurately. Figure \ref{fig:fourier} validates that weighted optimization results in better generalization in the overparameterized regime (Theorem \ref{thm:lower bound}), and shows non-degenerated double descent curves when $r=q=0.5$.

\paragraph{Function Interpolation} {\color{black}In this experiment, we interpolate the functions at $n$ equispaced points. The observed $n$ samples are $(\vx_j, y_j)_{j=1}^n$, where $y_j = f(\vx_j)$ with Fourier series $f(\vx) = \sum_{\vk} \theta_{\vk} \exp{(\pi i \vk^T\vx)}$. We fit training samples to a hypothesis class of $p$-truncated Fourier series: $ f_{\hat{\vtheta}}(\vx) = \sum_{\vk}  \hat{\theta}_{\vk} \exp{(\pi i \vk^T\vx)}$ with $k_j = -m, \dots, 0, \dots, m, \forall j \in [d]$ and then $p = (2m + 1)^d$ via least squares in underparameterized case; via plain and weighted min-norm estimators in the overparameterized case. $\mF_T^{\otimes d} \vtheta = {\bf y}$, where $\mF_T \in \mathbb{C}^{n_i \times p_i} \text{ with } F_{j,k} = e^{2 \pi i j k /n}$, $n = n_i^d$, and $p=p_i^d$. In the $1$-dimensional case, we use $D=1000$, $n=15$, $q = 1.5$ for $f_1(x)$ and $q=2$ for $f_2(x)$; in the $2$-dimensional case, $D=100^2$, $n=10^2$, $p_{under}=3^2$ and $p_{over}=41^2$.}
\begin{figure}[ht]
    \centering
    \includegraphics[width=\linewidth]{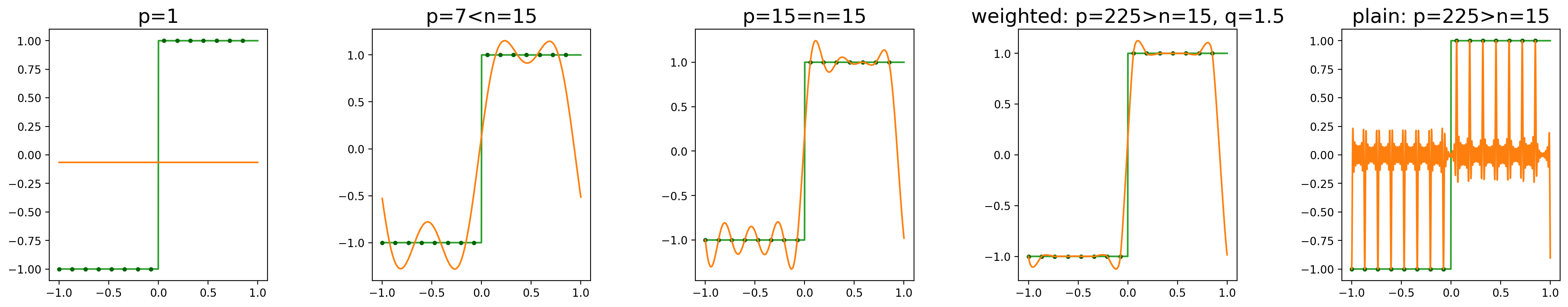}
    \includegraphics[width=\linewidth]{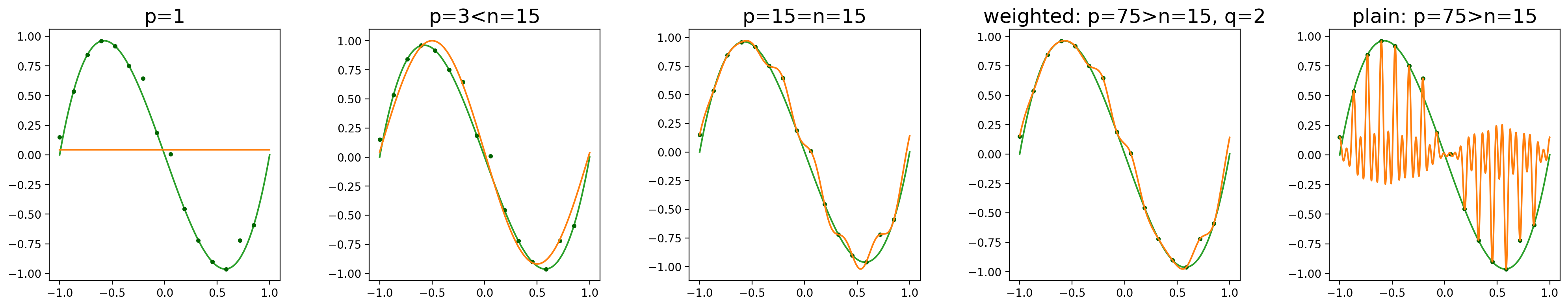}
    \vspace{-.6cm}
    \caption{Interpolation (in orange) of stage function and smooth function. Up: $f_1(x)=1, \forall x \in [-1,0); f_1(x) = 1, \forall x \in [0,1]$. Down: $f_2(x)= 2.5(x^3-x)$ with noise. From left to right: least square with $p=1$; least square with $p<n$; $p=n$; interpolation by weighted min-norm estimator ($p>n$); interpolation by plain min-norm estimator ($p>n$).} 
    \vspace{-.4cm}
    \label{fig:interp}
\end{figure}

Figure \ref{fig:interp} presents the function interpolations (in orange) using different estimators with the same set of equispaced samples (dark green). We observe that overparameterization with the plain min-norm estimator is useless, while the weighted min-norm estimator has the best performance in both noiseless and noisy cases. We also see the benefit of the weighted optimization's regularization towards smoother interpolants.

\begin{figure}[ht]
    \centering
    \includegraphics[width=.32\linewidth]{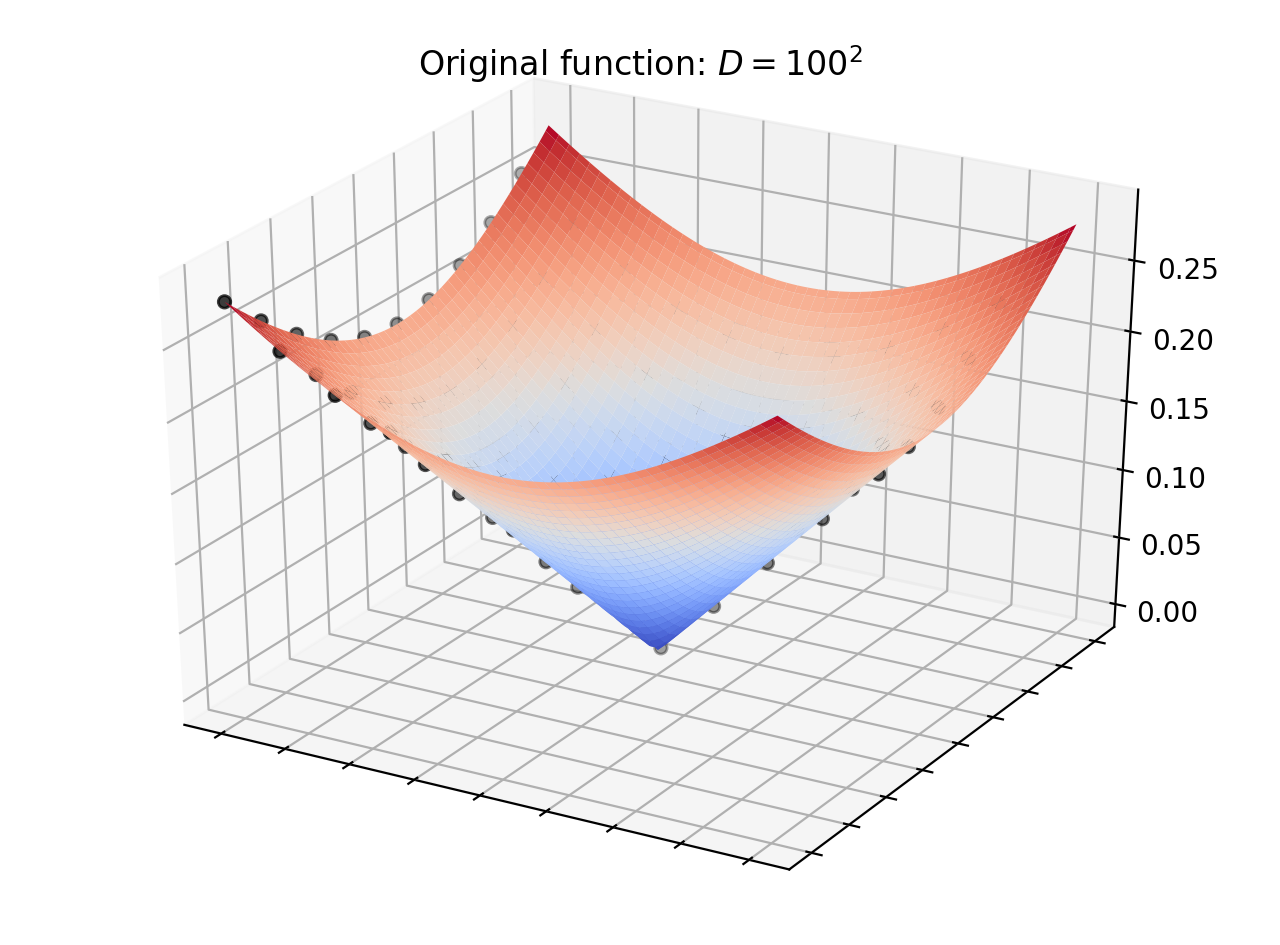}
    \includegraphics[width=.32\linewidth]{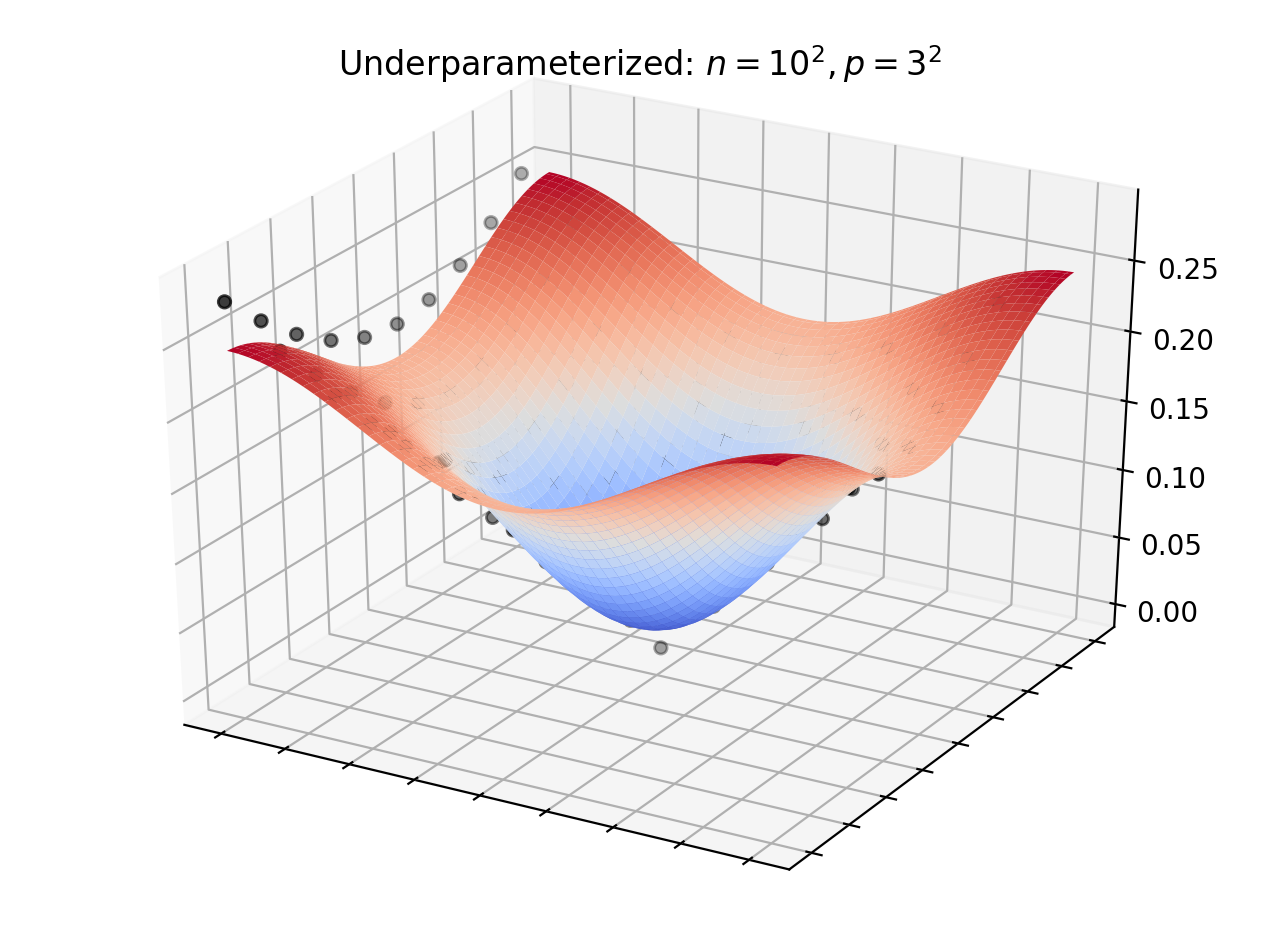}
    \includegraphics[width=.32\linewidth]{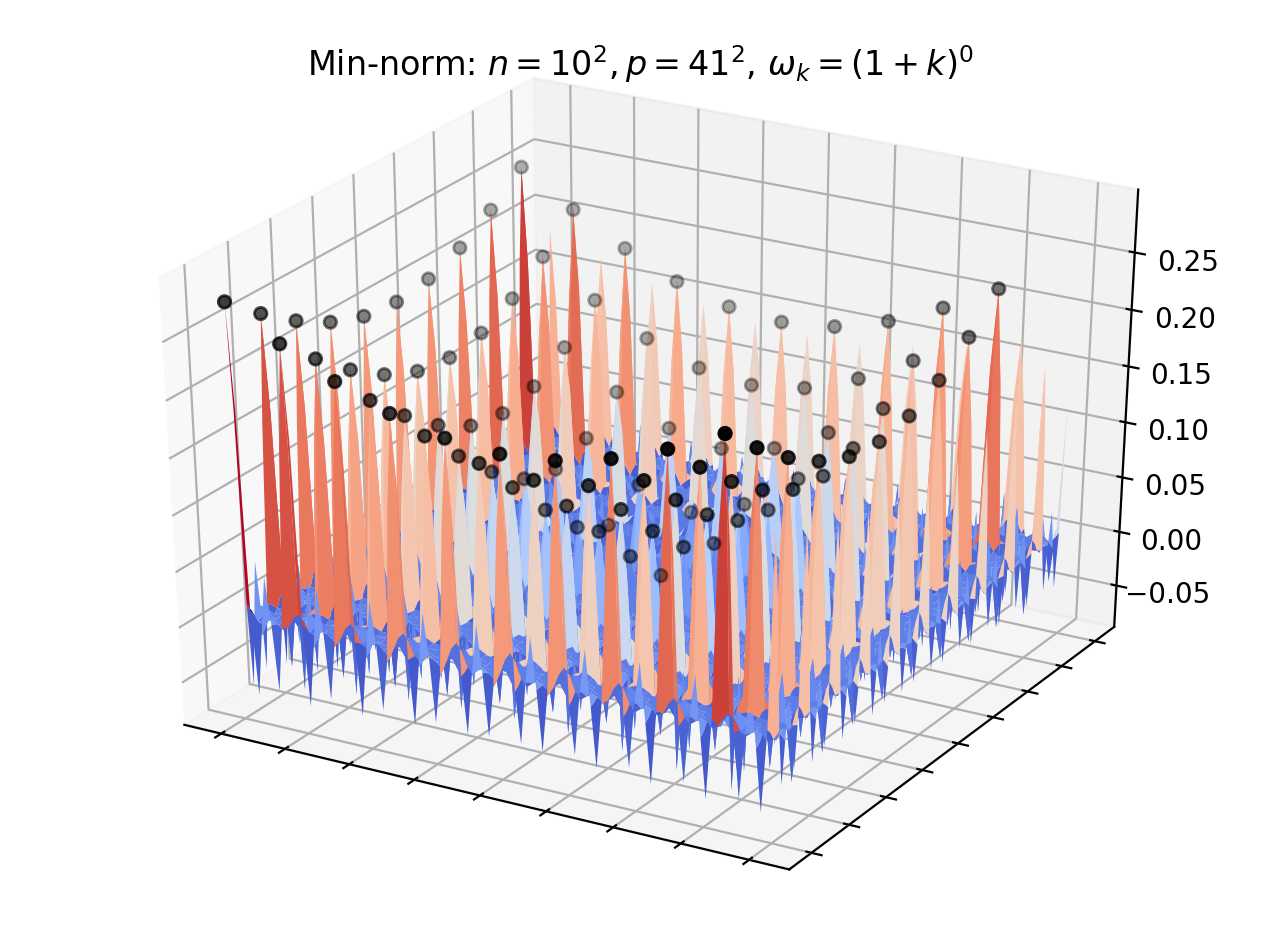}
    \includegraphics[width=.32\linewidth]{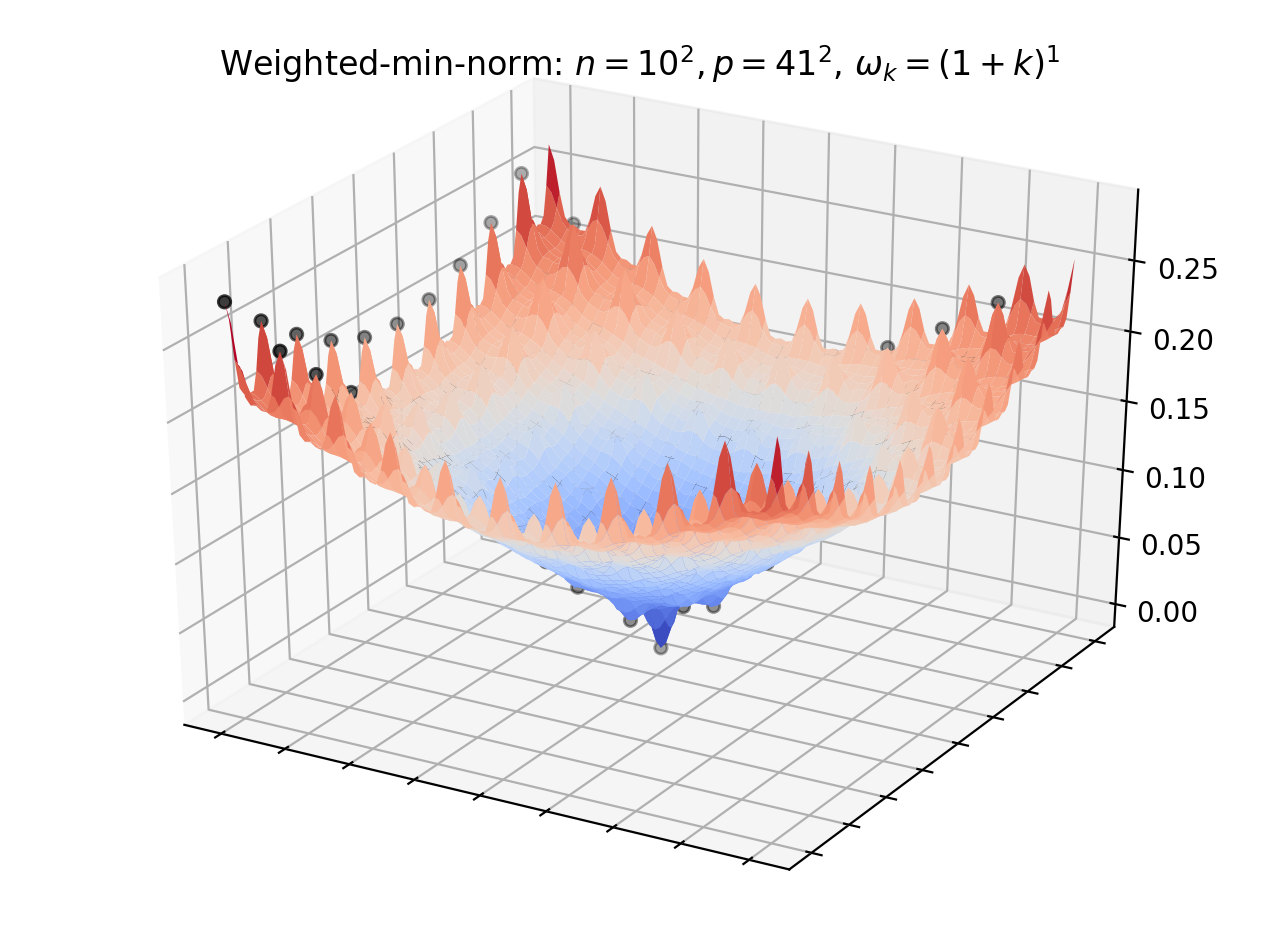}
    \includegraphics[width=.32\linewidth]{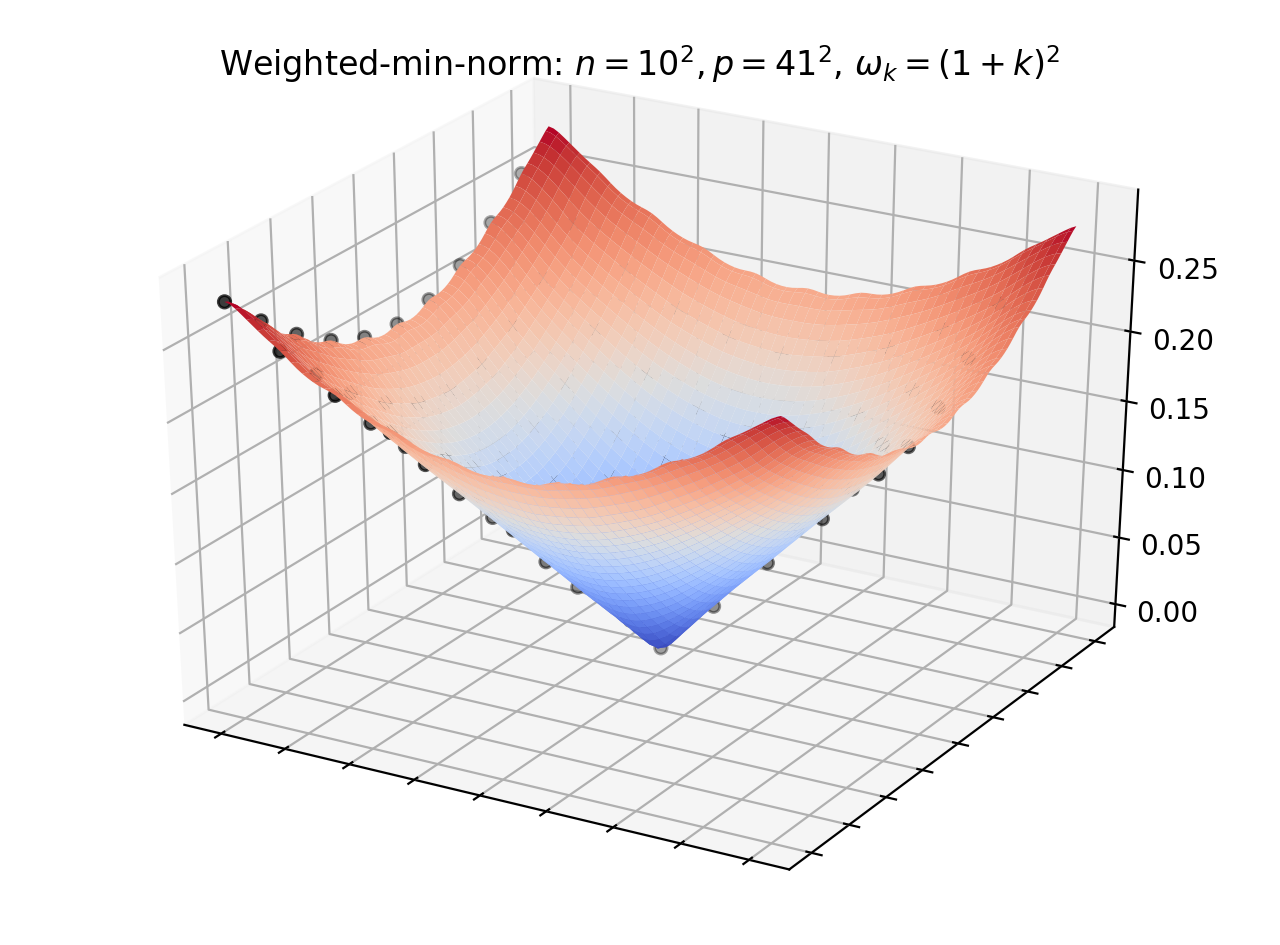}
    \includegraphics[width=.32\linewidth]{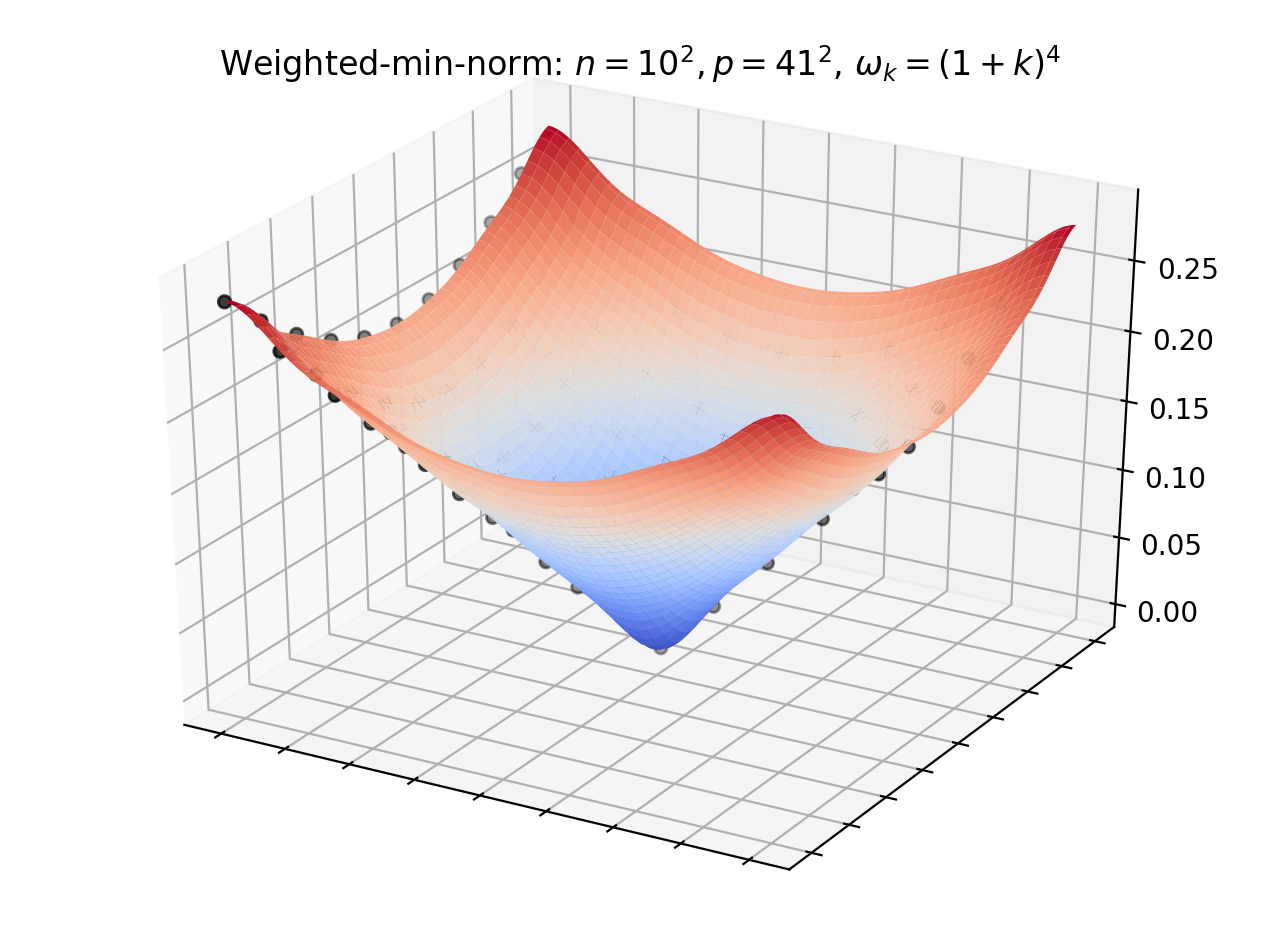}
    \vspace{-.4cm}
    \caption{Interpolation of $2D$ function $f(\vx) = \cos(6.28\cdot(2x+3y))$. Up: from left to right are original function with sampled points, least square with $p < n$, and interpolation with plain min-norm estimator ($p > n$); Bottom: from left to right are interpolations by weighted min-norm estimators ($p>n$) with $q=1, 2, 4$.} 
    \vspace{-.4cm}
    \label{fig:interp2d}
\end{figure}
{\color{black}
Figure \ref{fig:interp2d} shows a $2$-dimensional function interpolation. Similar to the $1$-dimensional case, the function approximated with $p<n$ is smooth but does not fit all the samples. The weighted min-norm estimator results in a smooth surface and almost recovers the original function with $q \geq 2$ while the plain min-norm estimator results in multiple sharp spikes.}

\section{Discussion on Necessity of Randomness}\label{sec:diss}
\begin{figure}[ht]
\centering
\vspace{-.2cm}
\includegraphics[width=0.46\textwidth]{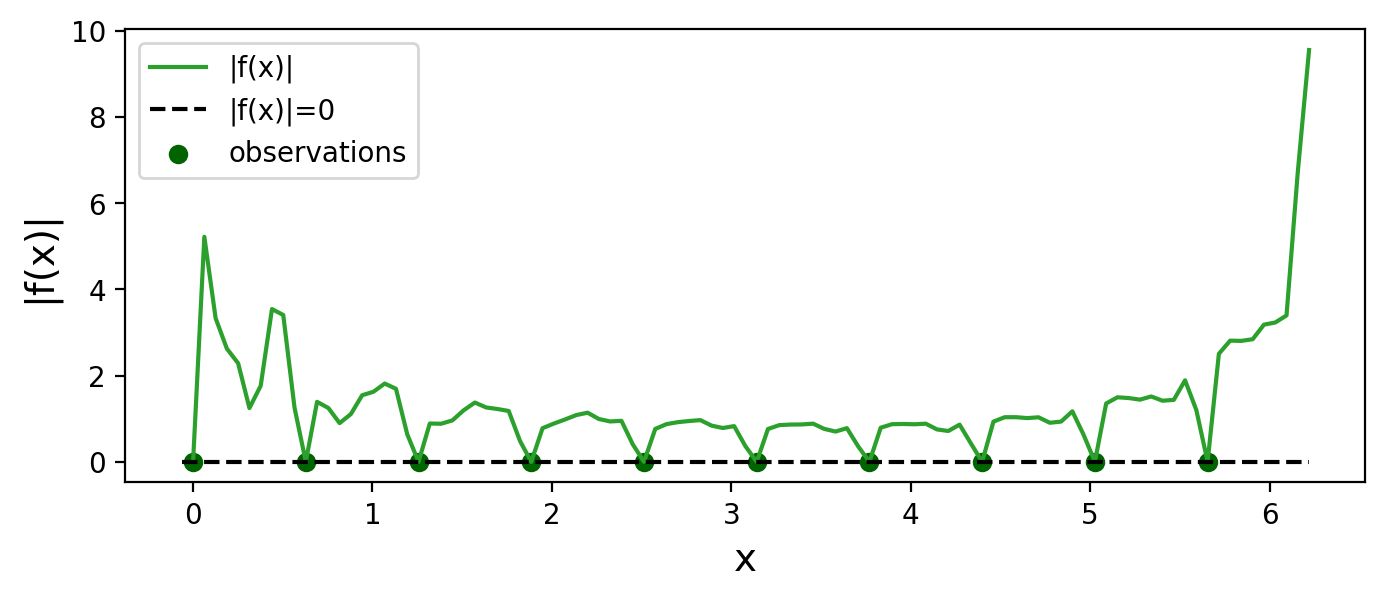}
\includegraphics[width=0.46\textwidth]{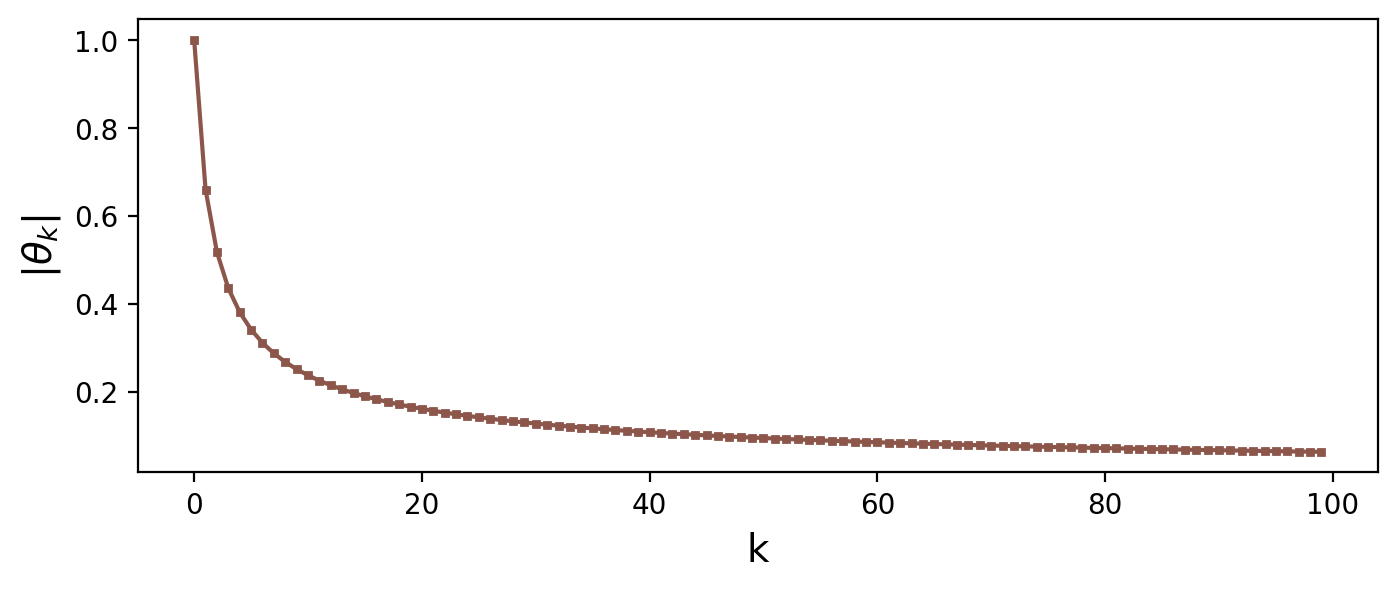}
\vspace{-.5cm}
\caption{In this example, we see that requiring $\theta_k$ decaying at rate $r$ is insufficient because it is possible that $f$ vanishes on all the sample points, and hence is impossible to recover. Here $D=100$, $n=10$, $r=0.6$.}
\label{fig:counter_example}
\vspace{-.1cm}
\end{figure}

In order to illustrate the necessity of randomness, we provide an example of a nontrivial function $f$ which has $r$-decaying Fourier coefficients and which vanishes at all the sample points $x_j$ in Figure~\ref{fig:counter_example}. According to the algorithm, the estimation via either plain or weighted min-norm optimization from such samples will be identically zero, which is clearly incorrect. The example is generated numerically by applying gradient descent to the loss function $L(\vphi):= \sum_{j=0}^{n-1} \big|\sum_{k=0}^{D-1} (k+1)^{-r}e^{i\phi_k} e^{\frac{2\pi i j k}{n}}\big|^2$. Once we find $\vphi^*$ such that $L(\vphi^*) = 0$, we set $f(x) = \sum_{k=0}^{D-1} (k+1)^{-r}e^{i\phi_k} e^{i k x}$, which vanished on all $x_j$ by construction.

In addition to the numerical experiment, we also prove the existence of such counterexamples. For $n=1$, finding $\vphi^* \st L(\vphi^*) = 0$ is reduced to constructing a $D$-polygon in $\mathbb{R}^2$ (equivalent to $\mathbb{C}$) with edges of length $(k+1)^{-r}$ for $k\in[D]$. For $r\leq 1$, since any edge is shorter than the sum of other edges, i.e., $(k'+1)^{-r}< \sum_{k\in[D]\backslash \{k'\}} (k+1)^{-r},\forall k'\in[D]$, such a polygon always exists, which can be proved by induction and the triangle inequality. Thus, there exists a nontrivial function which has $r$-decaying Fourier coefficients and vanishes at the origin ($x_0$).

\section{Conclusion and Outlook}\label{sec:sum}
This paper addresses an open question from Belkin, Hsu, and Xu  \cite{belkin2020two} on how and when weighted minimal $\ell_2$ norm trigonometric interpolation achieves low generalization error in the overparameterized regime. 
From our non-asymptotic expressions for the risk, we quantify how the bias towards smooth interpolations can be exploited to reduce the risk in the overparameterized setting and show that this risk is strictly better than the lowest possible risk in the underparameterized regime under certain conditions. In this way, our work also contributes to the understanding of the ``double descent" curve. 
 One particularly important direction for future research is to extend the setting of our sharp theoretical risk analysis beyond Fourier series models to general bounded orthonormal systems \cite{FR13} and neural networks. Another interesting direction of future work is to generalize our results to hold under a model of random sampling points rather than equidistant sampling points. Preliminary numerical experiments show similar behavior between equidistant and random sampling points, except around the pole $p=n$, where the risk for random samples blows up. We leave further investigations in this direction to future work.


\section*{Acknowledgments}
R. Ward and Y. Xie were supported in part by AFOSR 2018 MURI Award ``Verifiable, Control-Oriented Learning On The Fly". H.H Chou and H. Rauhut were supported in part by the DAAD grant 57417829 "Understanding stochastic gradient descent in  deep learning" and by the Excellence Initiative of the German federal and state governments.

\appendix
\section{Proof of Lemmas}\label{app:lemmas}

\begin{proof}[Proof of Lemma \ref{lem:weighted}] 
Using the re-parameterization $\vbeta =  \Sigma^{-q} \vtheta$ , the weighted min-norm estimator is $\hbeta_T := \tFst^{\dagger} \vy,  \hbeta_{T^c}:= \v0$, where $\vy = \tFst \bt + \tFstc \btc$ and $\tF = \mF \Sigma^q$. Since $\tFst$ has full rank, the
matrix $\tFst \tFst^* = \Fst \sTqs \Fst^*$ is invertible and $\tFst^{\dagger} = \tFst^*(\tFst \tFst^*)^{-1}$. Then,
\begin{align} \label{eq:risk_orig}
    \begin{split}
        \|\vtheta  - \hat{\vtheta}\|_2^2  & = \| \sTq (\bt -\hbeta_{T})\|^2 + {\color{black} \|\sTcq(\btc -\hbeta_{T^c})\|^2 } \\
        & = \| \sTq \bt - \sTq \dtFst ( \tFst \bt + \tFstc \btc) \|^2 + \|\sTcq \btc \|^2 \\
        & = \| \sTq  (\mI - \dtFst \tFst ) \bt - \sTq \dtFst \tFstc \btc \|^2 + \| \sTcq \btc\|^2 \\
        & =  \| \sTq (\mI - \dtFst\tFst ) \bt \|^2 + \| \sTq \dtFst \tFstc \btc\|^2 +  \|\sTcq \btc\|^2 
        \\ & \quad 
        -  \underbrace{ 
        2 \operatorname{Re} ( \bt ^* ( \mI - \dtFst \tFst ) \sTqs \dtFst \tFstc \btc ) }_{ =:\gC_1}.
    \end{split}
\end{align}
Since $ \dtFst\tFst$ is Hermitian, we have
\begin{align}\label{eq:proj}
        \| \sTq \p{ \mI - \dtFst\tFst } \bt \|^2 & = \norms{\sTq \bt} + \|\sTq \dtFst\tFst \bt \|^2  
         - \underbrace{
         2 ( \bt^* \sT^{2q} \dtFst\tFst \bt ) }_{=:\gC_2}.
\end{align}
Combining (\ref{eq:risk_orig}) and (\ref{eq:proj}) and taking expectation yields 
\begin{align} \label{eq:simple}
    \begin{split}
    \ex [ \| \Sigmaq (\vbeta -\hbeta) \|^2 ] & =  \ex [  \norms{\Sigmaq \vbeta} ] + \ex [ \| \sTq \dtFst\tFst \bt \|^2 ] +  \ex [ \|\sTq \dtFst \tFstc \btc\|^2 ]  -  \ex \bs{\gC_1}  -  \ex \bs{\gC_2} .
    \end{split}
\end{align}
The “trace trick” and $ \dtFst\tFst = \tFst^*(\tFst \tFst^*)^{-1}\tFst$ give 
\begin{align*}
    \begin{split}
         \ex  & {\color{black}[ \|  \sTq \dtFst\tFst \bt \|^2 ] }
         = \ex \bs{ \tr{\bt^* \dtFst\tFst \sTqs \dtFst\tFst \bt} } 
         = \tr{ \dtFst\tFst \sTqs \dtFst\tFst \ex \bs{ \bt \bt^* } } \\
        & = \tr{ {\color{black}\dtFst\tFst} \sTqs \dtFst \tFst \sT^{-q} \kt \sT^{-q} } 
         = {\color{black}\tr{ \sTqs \tFst^*(\tFst \tFst^*)^{-1}\Fst \kt \Fst^* (\tFst \tFst^*)^{-1}\tFst } }
    \end{split}
\end{align*}
Moreover,
\begin{align*}
    \begin{split}
    \ex [ \| \sTq \dtFst\tFstc \btc \|^2 ]
    & = \tr{ \tFstc^* (\dtFst)^* \sTqs \dtFst \tFstc \ex[ \btc \btc^* ] } \\
    & =  {\color{black} \tr{  \sTqs \dtFst \tFstc \sTc^{-q} \ktc \sTc^{-q}  \tFstc^* (\dtFst)^* } }   \\
    & = \tr{ \sTqs \tFst^*(\tFst \tFst^*)^{-1}\Fstc \ktc \Fstc^* (\tFst \tFst^*)^{-1}\tFst } .
    \end{split}
\end{align*}
Since $\ex \bs{ \btc \bt^*} = \sTc^{-q} \ex  \bs{ \vtheta_{T^c} \vtheta_T^*}  \sT^{-q} = 0$ we have $\ex \bs{\gC_1} = 0$. Furthermore, since $\mK$ commutes with $\Sigma^{-q}$ by diagonality, we have
$\sTqs \ex \left[ \bt \bt ^* \right] = \sTqs \ex \bs{ \Sigma^{-q} \vtheta \vtheta ^* \Sigma^{-q} } = \mK $ so that
\begin{align*}
    \begin{split}
        \ex \bs{\gC_2} & = 2 \, \tr{  \dtFst\tFst \sTqs \ex\left[ \bt \bt^*\right] } = 2\, \tr{ \kt  \dtFst\tFst} .  
    \end{split}
\end{align*}
Plugging all terms into (\ref{eq:simple}), we have
\begin{align*}
     \risk_q  & = \ex  [ \|  \Sigmaq (\vbeta -\hbeta)\|^2 ] \\
        &  = \tr{\mK }  +  \tr{ \Fst \sT^{4q} \Fst^*(\Fst \sTqs \Fst^*)^{-1}\Fstc \ktc \Fstc^* (\Fst \sTqs \Fst^*)^{-1} } \\ 
         & \quad  {\color{black} + \tr{ \Fst \sT^{4q} \Fst^*(\Fst \sTqs \Fst^*)^{-1}\Fst \kt \Fst^* (\Fst \sTqs \Fst^*)^{-1} } } \\
        & \quad {\color{black} - 2 \tr{ \Fst  \sTqs \kt \Fst^* (\Fst \sTqs \Fst^*)^{-1} } }.
\end{align*}
The risk of the plain min-norm estimator corresponds to $q=0$, which gives
\begin{align*}
       \risk_0 & = \ex  [  \| \vtheta - \hat{\vtheta}\|^2 ]  = \tr{\mK } - \tr{ \Fst \kt \Fst^* (\Fst \Fst^*)^{-1} }  +  \tr{\Fstc \ktc \Fstc^* (\Fst  \Fst^*)^{-1} } . 
\end{align*}
\end{proof}

\begin{proof} [Proof of Lemma \ref{lem:equi-fst}] 
For $ u \geq 0$, we set $ \mau = \Fst \stu \Fst^*$ and $\mcu = \Fstc \stcu \Fstc^*$, and define $\omn = \exp(-\frac{2\pi i}{n})$. Since $p=n l$, we have, for  $j_1, j_2 \in [n]$, 
\begin{align*}
\begin{split}
    (\mau)_{j_1,j_2} & = (\Fst \stu \Fst^*)_{j_1,j_2} = \sum_{k=0}^{p-1} t_k^u \exp \p{ \frac{-2\pi i}{n} (j_2 -j_1)  \cdot k }  
    = \sum_{\nu=0}^{l-1} \sum_{k=0}^{n-1} t_{k+n\nu}^u  \omn^{(j_2-j_1)k} ,\\
    (\mcu)_{j_1,j_2} & = (\Fstc \stcu \Fstc^*)_{j_1,j_2} =  \sum_{k=p}^{D-1} t_{k}^u \exp \p{\frac{-2\pi i}{n}  (j_2 -j_1)  \cdot k } 
    = \sum_{\nu=l}^{\tau -1}\sum_{k=0}^{n-1} t_{k+n\nu}^u \omn^{(j_2-j_1)k} .
\end{split}
\end{align*}
In the above equations, we use $\omn^{k+n\nu} = \omn^k$ for $\nu \in \sN_+$.

For $j \in [n]$, 
let $a_{j} = \aj $ and
$c_j = \cj $. Then 
\begin{align} \label{eq:prod}
\begin{split}
    (\mau)_{j_1,j_2} & =  \sum_{\nu=0}^{l-1} \sum_{k=0}^{n-1} t_{k+n\nu}^u  \omn^{(j_2-j_1)k} = a_{j_2-j_1 \pmod n} , \\ 
    (\mcu)_{j_1,j_2} & =   \sum_{\nu=l}^{\tau-1}\sum_{k=0}^{n-1} t_{k+n\nu}^u \omn^{(j_2-j_1)k}  = c_{j_2-j_1 \pmod n} .
\end{split}
\end{align}
Hence, for any $u \geq 0$, $\mau$ and $\mcu$ are circulant matrices.

For $u=0$, we use again $p = nl, l\in \sN_+$ to obtain that, for $j_1,j_2 \in [n]$,
\begin{align*}
  (\Fst & \Fst^*)_{j_1, j_2} =  \sum_{k=0}^{p-1}  \omn^{(j_2-j_1)k} =  
  \begin{cases}
   p, & \text{if} \quad j_1 = j_2, \\
    0  , & \text{if} \quad j_1 \neq j_2 .
  \end{cases}
\end{align*}
Hence, $\Fst \Fst^* = p\mI_n$ as claimed.
\end{proof}

\begin{proof}[Proof of Lemma~\ref{lem:sum_integral}]
By comparison to the sum to an integral, we have that
\begin{align*}
    \sum_{n=n_1}^{n_2}(an + b)^{-\alpha}
    &\geq \int_{n_1}^{n_2+1} (ax + b)^{-\alpha}dx
    = \frac{1}{a(\alpha-1)}((an_1 + b)^{-\alpha+1} - (a(n_2+1) + b)^{-\alpha+1})\\
    \sum_{n=n_1}^{n_2}(an + b)^{-\alpha}
    &= (an_1 + b)^{-\alpha} + \sum_{n=n_1+1}^{n_2}(an + b)^{-\alpha}
    \leq (an_1 + b)^{-\alpha} + \int_{n_1}^{n_2} (ax + b)^{-\alpha}dx\\
    &= (an_1 + b)^{-\alpha} + \frac{1}{a(\alpha-1)}((an_1 + b)^{-\alpha+1} - (an_2 + b)^{-\alpha+1}).
\end{align*}
In particular, for $\alpha=2r$, $a=1$, $b=0$, $n_1=1$, $n_2 = D$, 
\begin{align*}
    c_r^{-1}
    &\geq \frac{1}{2r - 1}(1 - (D+1)^{-2r+1})\\
    c_r^{-1}
    &\leq 1 + \frac{1}{2r-1}(1 - D^{-2r+1})
    = \frac{1}{2r-1}(2r - D^{-2r+1}).
\end{align*}
\end{proof}       

\begin{proof} [Proof of Lemma \ref{lem:ls}] 
In the under-parameterized setting, the error of the least squares estimator satisfies 
{\color{black}
\begin{align*}
   \| \vtheta - \hat{\vtheta}\|^2 &  = \norms{ (\Fst^* \Fst)^{-1} \Fst^* \p{ \Fst \bt + \Fstc \btc } - \bt } + \norms{ \btc } \\
    & = \norms{ (\Fst^* \Fst)^{-1} \Fst^*  \Fstc \btc }  + \norms{ \btc }  \\
    & =  \tr{ \Fstc^* \Fst (\Fst^* \Fst)^{-2} \Fst^* \Fstc \btc \btc^*   }   + \norms{ \btc } .
\end{align*} }
Taking expectation yields
\begin{align*}
    \ex  \bs{  \| \vtheta - \hat{\vtheta}\|^2  } =  \crs \tr{\sTcr} +  \crs  \tr{  \Fst (\Fst^* \Fst)^{-2} \Fst^* \Fstc \sTcr \Fstc^* } . 
\end{align*}
\end{proof}
\section{Proof of Theorems}\label{app:theorems}

\begin{proof}[Proof of Theorem~\ref{thm:order}]
To prove Theorem \ref{thm:order}, we start with a lemma that provides an explicit bound for the summation, and in particular $c_r$.
\begin{lemma}\textbf{(Bounds for the summation)}\label{lem:sum_integral}
For $n_1<n_2$ and $\alpha>1$,
\begin{align*}
    \sum_{n=n_1}^{n_2}(an + b)^{-\alpha}
    &\geq \frac{1}{a(\alpha-1)}((an_1 + b)^{-\alpha+1} - (a(n_2+1) + b)^{-\alpha+1})\\
    \sum_{n=n_1}^{n_2}(an + b)^{-\alpha}
    &\leq (an_1 + b)^{-\alpha} + \frac{1}{a(\alpha-1)}((an_1 + b)^{-\alpha+1} - (an_2 + b)^{-\alpha+1}).
\end{align*}
Consequently, for $r > 1/2$, the constant $\crs = (\sum_{j=0}^{D-1} (j+1)^{-2r})^{-1}$ satisfies
\begin{equation*}
    \frac{2r-1}{2r-D^{-2r+1}} \leq \crs \leq \frac{2r-1}{1-(D+1)^{-2r+1}}.
\end{equation*}
\end{lemma}

For $k \in [n]$ and $\alpha \in \sR$, we define 
\[
A(k,\alpha) := \sum_{\nu=0}^{l-1} t_{k+n\nu}^\alpha \quad \mbox{and} \quad B(k,\alpha) = \sum_{\nu=1}^{l-1} t_{k+n\nu}^\alpha = A(k,\alpha) - \frac{1}{(1+k)^\alpha},
 \]
 where we understand that $B(k,\alpha)=0$ if $l=1$.
 By Theorem~\ref{thm:diff-equi} we can write
 \begin{align*}
 1-\gP_q & = c_r\left(c_r^{-1} - \sum_{k=0}^{n-1}\frac{A(k,2q+2r)}{A(k,2q)}\right)\\
 & = c_r \sum_{k=0}^{n-1}\underbrace{\left(\frac{1}{(1+k)^{2r}}-\frac{A(k,2q+2r)}{A(k,2q)}\right)}_{=:\gamma_k} + c_r\sum_{k=n}^{D-1} \frac{1}{(1+k)^{2r}}.
 \end{align*}
 We have 
 \begin{align*}
 \gamma_{k-1} & = \frac{1}{k^{2r}} -  \frac{\frac{1}{k^{2q+2r}}+B(k-1,2q+2r)}{\frac{1}{k^{2q}}+B(k-1,2q)}
 = \frac{1}{k^{2r}} -  \frac{1+k^{2q+2r}B(k-1,2q+2r)}{k^{2r}+k^{2r+2q}B(k-1,2q)}\\
 & = \frac{1+k^{2q}B(k-1,2q)- 1 - k^{2q+2r}B(k-1,2q+2r)}{k^{2r}(1+k^{2q}B(k-1,2q))}\\
 & = \frac{k^{2q}B(k-1,2q)- k^{2q+2r}B(k-1,2q+2r))  }{k^{2r}(1+k^{2q}B(k-1,2q))} .
 \end{align*}
 Furthermore, if $l=1$ (i.e., $p=n$) then the numerator in the last expression vanishes and for $l > 1$ it satisfies
 \begin{align*}
 &k^{2q}B(k-1,2q)-k^{2q+2r}B(k-1,2q+2r)=\sum_{\nu=1}^{l-1} \left(\frac{k}{k+n\nu}\right)^{2q} -
 \sum_{\nu=1}^{l-1}\left(\frac{k}{k+n\nu}\right)^{2q+2r}\\
 &= \sum_{\nu=1}^{l-1}\left(\frac{k}{k+n\nu}\right)^{2q}\left(1 - \left(\frac{k}{k+n\nu}\right)^{2r}\right).
 \end{align*}
 Altogether, we have that $1- {\gP}_q = c_r \sum_{k=n+1}^{D} k^{-2r}$ if $l=1$ and for $l > 1$ it holds
 \begin{equation}\label{Pq-eq}
 1-{\gP}_q = c_r \sum_{\nu=1}^{l-1} \sum_{k=1}^n \frac{k^{2q-2r}}{1+k^{2q}B(k-1,2q)}  \frac{1}{(k+n\nu)^{2q}}\left(1-\left(\frac{k}{k+n\nu}\right)^{2r}\right) + c_r \sum_{k=n+1}^{D} \frac{1}{k^{2r}}.
 \end{equation}
For $r > 1/2$, the last term can be upper bounded by $c_r(2r-1)^{-1} n^{-2r+1}$ according to lemma \ref{lem:sum_integral}. Similarly,
\begin{align*}
    B(k-1,2q) &\geq \frac{1}{n(2q-1)}\left((k+n)^{-2q+1} - (k+\ell n)^{-2q+1}\right),
\end{align*}
where we use the fact that $q>1/2$. Since the last expression is decreasing with $k \in \{1,\hdots,n\}$, we obtain the lower bound
 \begin{align*}
 B(k-1,2q)& \geq \frac{1}{n(2q-1)}\left( (2n)^{-2q+1} - ((l+1)n)^{-2q+1}\right) = \frac{d_q}{n^{2q}}, 
 \end{align*}
where $d_q:= \frac{2^{-2q+1}-(l+1)^{-2q+1}}{2q-1}$.
 Hence, we have
 \begin{align*}
 1-\gP_q & \leq \frac{c_r}{1+d_q n^{-2q}} \sum_{\nu=1}^{l-1} \sum_{k=1}^n  \frac{k^{2q-2r}}{(k+n\nu)^{2q}}\left(1-\left(\frac{k}{k+n\nu}\right)^{2r}\right) + \frac{c_r}{2r-1} n^{-2r+1}\\
 &\leq \frac{c_r}{1+d_q n^{-2q}} \sum_{\nu=1}^{l-1} \sum_{k=1}^n  \frac{k^{2q-2r}}{(k+n\nu)^{2q}} + \frac{c_r}{2r-1} n^{-2r+1}.
 \end{align*}
 If $q=r$ then the double sum above can be estimated as
 \[
\sum_{\nu=1}^{l-1}\sum_{k=1}^n \frac{1}{(k+n\nu)^{2q}} = \sum_{j=n+1}^p \frac{1}{j^{2q}} 
\leq 
\int_n^p \frac{1}{x^{2q}} dx = \frac{1}{2q-1}\left(n^{-2q+1} - p^{-2q+1}\right).
 \]
Altogether, for $r=q > 1/2$ and $p = ln$ for $l\geq 2$,
 \begin{align*}
 1 - \gP_q & \leq \frac{c_r}{2r-1}\left(\frac{n^{-2r+1}-p^{-2r+1}}{1+d_r n^{-2r}} + n^{-2r+1}\right) \\
 & \leq \frac{1}{1-(D+1)^{-2r+1}}\left(\frac{n^{-2r+1}-p^{-2r+1}}{1+d_r n^{-2r}} + n^{-2r+1}\right),
 \end{align*}
 where we have used Lemma~\ref{lem:sum_integral} in the last step. 
 
It remains to bound {\color{black}$\gQ_{q,1}$ and  $\gQ_{q,2}$} from above.
Towards this goal, we observe the simple inequality
 $\sum_{\nu=0}^{l-1} t_{k+n\nu}^{4q} \leq \left(\sum_{\nu=0}^{l-1} t_{k+n \nu}^{2q} \right)^2.
 $
Thus by Lemma \ref{lem:sum_integral}, we have the immediate bound 
 \begin{align*}
 {\color{black}\gQ_{q,2}} &\leq c_r\sum_{k=0}^{n-1} \sum_{\nu = l}^{\tau - 1} t_{k+n \nu}^{2r} = c_r \sum_{k=p+1}^{D} \frac{1}{k^{2r}}
 \leq \frac{p^{-2r+1}}{1 - (D+1)^{-2r+1}}.
 \end{align*}
 
{ \color{black}
 For $\gQ_{q,1}$, if $q=r$, we have
 \begin{align*}
-\gP_q & + \gQ_{q, 1}  =  -  \crs \sum_{k=0}^{n-1} \frac{ \sinsuma{ {2q+2r} } }{ \sinsuma{ {2q} } }   + 
 \crs \sum_{k=0}^{n-1} \frac{ \p{ \sinsuma{ {4q} } } \p{ {\color{black} \sum_{\nu=0}^{l-1} } t_{k+n\nu}^{2r} }  }{ \p{ \sinsuma{ {2q} } } ^2 }  \\
 & = \crs \sum_{k=0}^{n-1} [ - ( \sinsuma{ {2r+2r} } ) ( \sinsuma{ {2r} } )   + ( \sinsuma{ {4r} } ) ( \sum_{\nu=0}^{l-1}  t_{k+n\nu}^{2r} ) ]/ ( \sinsuma{ {2r} } )^2   = 0
 \end{align*}
 }

 Altogether, for $r=q > 1/2$ and $p=ln$ with $l\geq 2$,
 \begin{align}
 \risk_q & = 
 1 - {\color{black} 2\gP_q +  \gQ_{q,1} + \gQ_{q,2} = 1-  \gP_q + \gQ_{q, 2} + (\gQ_{q,1} - \gP_q ) } \\
 & \leq \frac{1}{1-(D+1)^{-2r+1}}\left(\left(\frac{1}{1+d_rn^{-2r}}+1\right)n^{-2r+1} + \left(1-\frac{1}{1+d_r n^{-2r}}\right) p^{-2r+1}\right)\notag\\
 & = \frac{1}{1-(D+1)^{-2r+1}}
 \left(\frac{2+d_r n^{-2r}}{1+d_r n^{-2r}} n^{-2r+1} + \frac{d_r}{1+d_r n^{-2r}} n^{-2r} p^{-2r+1}
 \right).\label{bound:risk}
 \end{align}
 The previous expression can be bounded by $Cn^{-2r+1}$ for a suitable constant $C$.
 
 If $l=1$ so that $p=n$ then the above derivations give
 \[
 \risk_q = 1 - {\color{black} 2 \gP_q +  \gQ_{q,1} + \gQ_{q,2} }
 \leq \frac{1}{1-(D+1)^{-2r+1}}\left(n^{-2r+1} + p^{-2r+1}\right) = \frac{2 n^{-2r+1}}{1-(D+1)^{-2r+1}} ,
 \]
 which gives the statement of the theorem also in this case.
 \end{proof}

\begin{proof}[Proof of Theorem \ref{thm:highprob}]
From Lemma \ref{lem:weighted}, we have 
\begin{equation}\label{eq1}
    \|\vtheta  - \hat{\vtheta}\|_2^2 
    = \norms{\sTq \left(\mI - \dtFst\tFst \right) \bt }  + \norms{\sTq \dtFst \tFstc \btc} +  \norms{\sTcq \btc} - \gC_1,
\end{equation}
where $\gC_1 = 2 \operatorname{Re} \left( \bt ^* \left( \mI - \dtFst \tFst \right) \sTqs \dtFst \tFstc \btc \right)$. Equation (\ref{eq1}) can be expressed as 
\begin{equation*}
    \|\vtheta  - \hat{\vtheta}\|_2^2 
    = \begin{bmatrix}\bt^* & \btc^* \end{bmatrix}
    \begin{bmatrix}
    D_1 & B\\
    B^* & D_2
    \end{bmatrix}
    \begin{bmatrix}\bt \\ \btc \end{bmatrix}
    = \vbeta^* M \vbeta,
\end{equation*}
where $D_1 = \left(\mI - \dtFst\tFst \right)\sTqs \left(\mI - \dtFst\tFst \right)$, $D_2 = \Sigma_{T^c}^{2q} + \tFstc^* (\dtFst)^* \sTqs \dtFst \tFstc$, 
and $B = \left( \mI - \dtFst \tFst \right) \sTqs \dtFst \tFstc$.
Since $\vtheta$ (and hence $\vbeta$) has independent sub-Gaussian coordinates, the Hanson-Wright inequality gives
\begin{align*}
    \mathbb{P}(|\vbeta^* M \vbeta - \mathbb{E}\vbeta^* M \vbeta| \geq t)
    &\leq 2\text{exp}\left[-c\min\left(\frac{t^2}{K^4\|M\|_F^2}, \frac{t}{K^2\|M\|}\right)\right],
\end{align*}
where $K = \max_k\|\vbeta_k\|_{\psi_2} = \sqrt{c_r}$ because $r\geq q$. Since $\|M\|_2\leq \|M\|_F$, it suffices to bound
\begin{equation*}
    \|M\|_F^2=\text{tr}(D_1^*D_1 + 2B^*B + D_2^*D_2).
\end{equation*}
Since $\tFst$ has full rank, $\tFst^{\dagger} = \tFst^*(\tFst \tFst^*)^{-1} = \sTq\Fst^* (\Fst \sTqs \Fst^*)^{-1} $. By the circulant property from Lemma \ref{lem:equi-fst}, we have
\begin{align*}
    \text{tr}(\tFst \stu \dtFst) 
    &= \text{tr}(\Fst \sTq \stu \sTq\Fst^* (\Fst \sTqs \Fst^*)^{-1}) 
    = \text{tr}(\Lambda_{a,2q+u}\Lambda_{a,2q}^{-1})\\
    \text{tr}((\dtFst)^* \stu \dtFst)
    &= \text{tr}(( (\Fst \sTqs \Fst^*)^{-1})^*\Fst\sTq \stu \sTq\Fst^* (\Fst \sTqs \Fst^*)^{-1})
    = \text{tr}(\Lambda_{a,2q+u}\Lambda_{a,2q}^{-2}).
\end{align*}
An explicit calculation yields
\begin{align*}
        \text{tr}\left(D_1^*D_1\right)
        &= \text{tr}\left( (\mI - \dtFst\tFst )\sTqs (\mI - \dtFst\tFst )^2\sTqs (\mI - \dtFst\tFst )\right)\\ 
        &= \text{tr}\left( \left[\sTqs (\mI - \dtFst\tFst )^2\right]^2 \right)
        = \text{tr}\left( \left[\sTqs (\mI -\dtFst\tFst)\right]^2 \right)\\
        &= \text{tr}\left(\sT^{4q} - 2 \tFst \sT^{4q}\dtFst + (\tFst \sTqs\dtFst)^2\right)
        = \text{tr}\left(\sT^{4q} -2\Lambda_{a,6q}\Lambda_{a,2q}^{-1} + (\Lambda_{a,4q}\Lambda_{a,2q}^{-1})^2 \right);\\
        \text{tr}(D_2^*D_2)
        &= \text{tr}\left(\Sigma_{T^c}^{4q} + 2\Sigma_{T^c}^{2q}\tFstc^* (\dtFst)^* \sTqs \dtFst \tFstc + \tFstc^* (\dtFst)^* \sTqs \dtFst \tFstc \tFstc^* (\dtFst)^* \sTqs \dtFst \tFstc \right)\\
        &= \text{tr}\left(\Sigma_{T^c}^{4q} + 2 \tFstc\Sigma_{T^c}^{2q}\tFstc^* (\dtFst)^* \sTqs \dtFst + (\tFstc\tFstc^*(\dtFst)^* \sTqs \dtFst )^2  \right)\\
        &= \text{tr}\left(\Sigma_{T^c}^{4q} + 2 \Lambda_{c, 4q} \Lambda_{a,4q}\Lambda_{a,2q}^{-2} + (\Lambda_{c, 2q} \Lambda_{a,4q}\Lambda_{a,2q}^{-2})^2  \right);\\
        \text{tr}(B^*B)
        &= \text{tr}\left( \tFstc^*(\dtFst)^*\sTqs (\mI - \dtFst \tFst )^2 \sTqs \dtFst \tFstc \right)\\
        &= \text{tr}\left( \tFstc\tFstc^*(\dtFst)^*\sTqs (\mI - \dtFst \tFst ) \sTqs \dtFst  \right)\\
        &= \text{tr}\left( \tFstc\tFstc^*(\dtFst)^*\sT^{4q} \dtFst -\tFstc\tFstc^*(\dtFst)^*\sTqs \dtFst \tFst \sTqs \dtFst \right)\\
        &= \text{tr}\left( \Lambda_{c, 2q}\Lambda_{a,6q}\Lambda_{a,2q}^{-2} - \Lambda_{c, 2q}\Lambda_{a,4q}\Lambda_{a,2q}^{-2}\Lambda_{a,4q}\Lambda_{a,2q}^{-1} \right).
\end{align*}
Therefore,
\begin{align}\label{eq:MF}
    \begin{split}
        \|M\|_F^2
        &= \text{tr}\left(D_1^*D_1+D_2^*D_2+BB^*+B^*B\right)\\
        &= \text{tr}\left(\Sigma^{4q}+(\Lambda_{a,4q}^2 +2\Lambda_{c,4q}\Lambda_{a,4q} +2\Lambda_{c,2q}\Lambda_{a,6q})\Lambda_{a,2q}^{-2} +\Lambda_{c,2q}^2\Lambda_{a,4q}^2\Lambda_{a,2q}^{-4}\right)\\
        &\quad -\text{tr}\left(2\Lambda_{a,6q}\Lambda_{a,2q}^{-1} +2\Lambda_{c,2q}\Lambda_{a,4q}^2\Lambda_{a,2q}^{-3}\right).
    \end{split}
\end{align}
We will now bound this expression using the information on $\Lambda$. Note that $\lambda_{a,q}^{(s)}$, the $s$-th diagonal entry of $\Lambda_{a,q}$, follows the inequality $\lambda_{a,q_1+q_2}^{(s)}\leq \lambda_{a,q_1}^{(s)}\lambda_{a,q_2}^{(s)}$ for any $q_1,q_2>0$. Hence, 
{\color{black}
\begin{align*}
    \text{tr}\left(\Lambda_{a,4q}^2 \Lambda_{a,2q}^{-2}\right)
    &\leq \text{tr}\left(\Lambda_{a,4q}\Lambda_{a,2q}^2 \Lambda_{a,2q}^{-2}\right)
    = \text{tr}\left(\Lambda_{a,4q}\right)
    = \text{tr}(\Sigma_T^{4q})\\
    \text{tr}\left(\Lambda_{c,4q}\Lambda_{a,4q}\Lambda_{a,2q}^{-2}\right)
    &\leq \text{tr}\left(\Lambda_{c,4q}\Lambda_{a,2q}^2 \Lambda_{a,2q}^{-2}\right)
    = \text{tr}\left(\Lambda_{c,4q}\right)
    = \text{tr}(\Sigma_{T^c}^{4q})\\
    \text{tr}\left(\Lambda_{c,2q}\Lambda_{a,6q}\Lambda_{a,2q}^{-2}\right)
    &\leq \text{tr}\left(\Lambda_{c,2q}\Lambda_{a,2q}^3\Lambda_{a,2q}^{-2}\right)
    = \text{tr}\left(\Lambda_{c,2q}\Lambda_{a,2q}\right)
    \leq \text{tr}(\Sigma_T^{2q})\text{tr}(\Sigma_{T^c}^{2q})\\
    \text{tr}\left(\Lambda_{c,2q}^2\Lambda_{a,4q}^2\Lambda_{a,2q}^{-4}\right)
    &\leq \text{tr}\left(\Lambda_{c,2q}^2\Lambda_{a,2q}^4\Lambda_{a,2q}^{-4}\right)
    = \text{tr}\left(\Lambda_{c,2q}^2\right)
    \leq \text{tr}(\Sigma_{T^c}^{2q})^2
\end{align*} 
}
Lemma \ref{lem:sum_integral} implies that for $\alpha>1$,
\begin{align*}
    \frac{1}{2(\alpha-1)}
    \leq\text{tr}(\Sigma^\alpha)
    \leq \frac{\alpha}{\alpha-1},
    \quad K^2 \leq 2(2r - 1),
\end{align*}
and hence,
\begin{align*}
    \|M\|_F^2
    &\leq \text{tr}\left(\Sigma^{4q} +\Sigma_T^{4q} +2\Sigma_{T^c}^{4q} \right)
    +2\text{tr}(\Sigma_T^{2q}) \text{tr}(\Sigma_{T^c}^{2q})
    +\text{tr}(\Sigma_{T^c}^{2q})^2\\
    &\leq \frac{12q}{4q-1} + \frac{16q^2}{(4q-1)(2q-1)} + \frac{4q^2}{(2q-1)^2}
    = \frac{4q(24q^2 - 17q + 3)}{(2q-1)^2(4q-1)}.
\end{align*}
The conclusion then follows by plugging all the bounds into the Hanson-Wright inequality.
\end{proof}

\begin{proof} [Proof of Theorem \ref{thm:lower bound}] 
(a) We show that, for fixed $n$ and $r\geq 1$, the risk in the underparameterized setting is monotonically decreasing in $p$. To this end, we set $f(p) = \sum_{j=p}^{D-1} t_j^{2r} + \sum_{k=1}^{\tau-1} \sum_{j=0}^{p-1} t_{kn+j}^{2r}$, where $t_j = (j+1)^{-1}$. Let $\Delta(p) = f(p+1) - f(p)$ be the increment. Then $\Delta(p) = - t_{p}^{2r} + \sum_{k=1}^{\tau-1} t_{kn+p}^{2r}$. The goal is to show $\Delta(p)\leq 0$ for all $p\in[n]$. Since $2r>1$, by Lemma \ref{lem:sum_integral} 
\begin{align*}
\vspace{-.5cm}
    \Delta(p) 
    &\leq - (p + 1)^{-2r} + \frac{1}{n(2r-1)}[(p + 1)^{-2r+1} - (n(\tau-1) +p + 1)^{-2r+1}]
    =: \Delta^+(p).
\end{align*}
Hence, it suffices to show that $\Delta^+(p)\leq 0$ for $p\in[n]$. Its derivative w.r.t. $p$ is
\begin{align*}
    {\Delta^+}'(p)
    &= 2r (p + 1)^{-2r-1} + \frac{1}{n}[(n(\tau-1) +p + 1)^{-2r} - (p + 1)^{-2r}]\\
    &= (p + 1)^{-2r}\left(2r(p+1)^{-1} - \frac{1}{n}\right)+ \frac{1}{n}(n(\tau-1) +p + 1)^{-2r}
    >0
\end{align*}
since $2nr > p+1$ and hence all the terms are positive. Because $\Delta^+$ is increasing, it suffices to check if the end point, $\Delta^+(n-1)$, is non-positive to ensure $\Delta^+\leq 0$. Indeed,
\vspace{-.2cm}
\begin{align*}
    \Delta^+(n-1)
    &= - n ^{-2r} + \frac{1}{n(1-2r)}[(n\tau)^{1-2r} - n^{1-2r}]
    = -n ^{-2r} + \frac{n^{-2r}}{2r-1}[1 - \tau^{1-2r}]\\
    &= -n^{-2r}\left(1 - \frac{1}{2r-1} \right) - \frac{n^{-2r}\tau^{1-2r}}{2r-1} <0
\end{align*}
since $2r - 1 >1$. Then the lowest risk is
$\risk_{under}^* = 2 \crs \sum_{j=n}^{D-1} t_j^{2r}$.
\\
(b) Next, we consider two cases in the overparameterized regime: $p=n$ and $p=D$ for $q=r$ first. When $p=n$ (i.e., $l=1$), the risk can be written as 
\begin{align*}
        \risk_q (p=n) & = 1 - \crs \sum_{k=0}^{n-1} t_k^{2r} + \crs \sum_{k=0}^{n-1} \sum_{\nu=l}^{\tau -1 }t_{k+n\nu}^{2r} 
         = 1 - \crs \sum_{k=0}^{n-1} ( t_{k}^{2r} -  \sum_{\nu=1}^{\tau-1} t_{k+n\nu}^{2r}) := 1-\crs\sum_{k=0}^{n-1} b_k.
\end{align*}
When $p=D$ (i.e., $l=\tau$) {\color{black} and $q=r$, } 
\begin{align*}
        \risk_r (p=D) & = {\color{black} 1 -  2 \crs \sum_{k=0}^{n-1} \frac{  \sum_{\nu=0}^{\tau -1} t_{k+n\nu}^{2r+2r}  }{  \sum_{\nu=0}^{\tau-1} t_{k+n\nu}^{2r}  }   + \crs \sum_{k=0}^{n-1}  \frac{ \p{ \sum_{\nu=0}^{\tau -1} t_{k+n\nu}^{4r} } \p{ \sum_{\nu=0}^{\tau -1} t_{k+n\nu}^{2r} }  }{ \p{ \sum_{\nu=0}^{\tau-1} t_{k+n\nu}^{2r} } ^2 } }  \\
        & = 1 - \crs \sum_{k=0}^{n-1} \frac{ \sinsuma{ {4r} } }{ \sinsuma{ {2r} } }  
        = 1 - \crs \sum_{k=0}^{n-1} \frac{ \sum_{\nu=0}^{\tau -1} t_k^{4r}  }{ t_{k}^{2r} + \sum_{\nu=1}^{\tau-1} t_{k+n\nu}^{2r} } := 1-\crs\sum_{k=0}^{n-1}d_k . 
\end{align*}
The quotient $d_k/b_k$ satisfies
{\color{black}
\begin{align}\label{eq:quotient}
    \begin{split}
       \frac{d_k}{b_k} & = \frac{t_k^{4r} + \sum_{\nu=1}^{\tau-1} t_{k+n\nu}^{4r}  }{ \p{ t_{k}^{2r} + \sum_{\nu=1}^{\tau-1} t_{k+n\nu}^{2r} }  \p{ t_{k}^{2r} -  \sum_{v=1}^{\tau-1} t_{k+n\nu}^{2r} } }  
       = \frac{t_k^{4r} + \sum_{\nu=1}^{\tau-1} t_{k+n\nu}^{4r}  }{  t_{k}^{4r} -  \p{ \sum_{v=1}^{\tau-1} t_{k+n\nu}^{2r} } ^2 } .
    \end{split}
\end{align} 
If $b_k := t_k^{2r} - \sum_{\nu=1}^{\tau -1} t_{k+n \nu}^{2r} \leq 0$, $\risk_r(p=D) \leq \risk_r(p=n)$ holds obviously; if $b_k >0$, $d_k/b_k \geq 1$ from \eqref{eq:quotient}. Hence, for $q=r$, we have $\risk_r(p=D) \leq \risk_r(p=n)$.
\\
Then, consider the case $q >r$. For $p \geq n$, define $z_p= \frac{ 2\sinsuma{ {2q+2r} } }{ \sinsuma{ {2q} } } - \frac{ \p{ \sinsuma{ {4q} } } \p{  \sum_{\nu=0}^{\tau-1}  t_{k+n\nu}^{2r} }  }{ \p{ \sinsuma{ {2q} } } ^2 } $, then $\risk_q(p) = 1 - \crs z_p$.
\begin{align*}
    z_{2n} & := \frac{2(t_k^{4q+2r} + t_{k+n}^{4q+2r} + t_k^{2q+2r}t_{k+n}^{2q} +t_k^{2q}t_{k+n}^{2q+2r}) - (t_k^{4q} +  t_{k+n}^{4q} )  \sum_{\nu=0}^{\tau-1}  t_{k+n\nu}^{2r} }{t_k^{4q} + t_{k+n}^{4q} + 2t_k^{2q}t_{k+n}^{2q}} := \frac{z_{2n, u}}{z_{2n, d}} \\
    z_n & := \frac{2(t_k^{4q+2r} + t_k^{2r}t_{k+n}^{4q} + 2 t_k^{2q+2r}t_{k+n}^{2q}) - (t_k^{4q} + t_{k+n}^{4q} + 2t_k^{2q}t_{k+n}^{2q})  \sum_{\nu=0}^{\tau-1}  t_{k+n\nu}^{2r} }{t_k^{4q} + t_{k+n}^{4q} + 2t_k^{2q}t_{k+n}^{2q}} := \frac{z_{n, u}}{z_{n, d}} \\
   z_{n, u} & - z_{2n, u} =  2(t_k^{2r} t_{k+n}^{4q} + t_k^{2q+2r}t_{k+n}^{2q} - t_{k+n}^{4q+2r} -  t_k^{2q}t_{k+n}^{2q+2r} ) - 2t_k^{2q}t_{k+n}^{2q} (t_k^{2r}  + \text{res}) \\
   & = 2(t_k^{2r} t_{k+n}^{4q} - t_k^{2q}t_{k+n}^{2q+2r} - t_{k+n}^{4q+2r} ) - 2t_k^{2q}t_{k+n}^{2q} \text{res}
\end{align*}
For $q \geq r > 0$, we have $\frac{ t_k^{2q}t_{k+n}^{2q+2r} }{t_k^{2r} t_{k+n}^{4q} } = (\frac{t_k}{t_{k+n}})^{2q-2r} = (\frac{k+n+1}{k+1})^{2q-2r} \geq 1$, then $t_k^{2r} t_{k+n}^{4q} \leq t_k^{2q}t_{k+n}^{2q+2r}$. Hence $z_{n,u} \leq z_{2n, u}$, and then $\risk_q(p=2n) \leq \risk_q(p=n) = \risk^*_{under} , \forall q\geq r \geq 1 $, i.e., the lowest risk in the over- regime is strictly less than that in the underparameterized regime.
}
\end{proof}
\section{Visualization of Theoretical Risks} \label{exp:other}
\begin{figure}[ht]
    \centering
    \includegraphics[width=.28\linewidth]{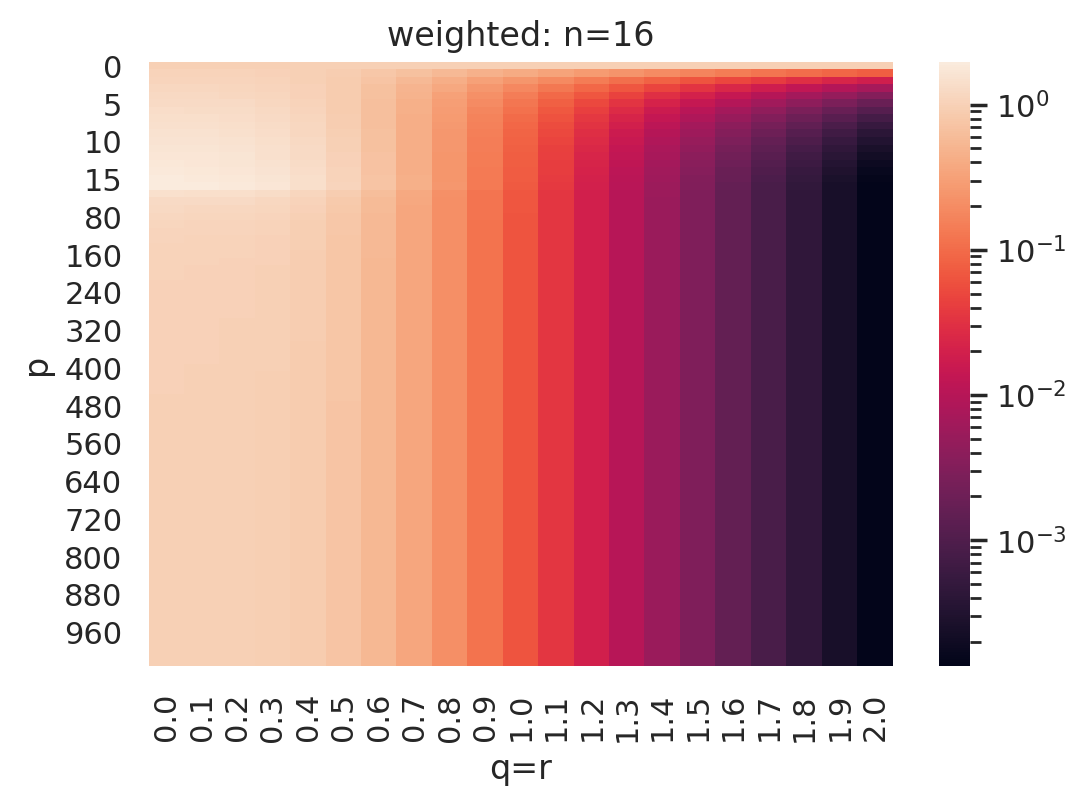}
    \includegraphics[width=.28\linewidth]{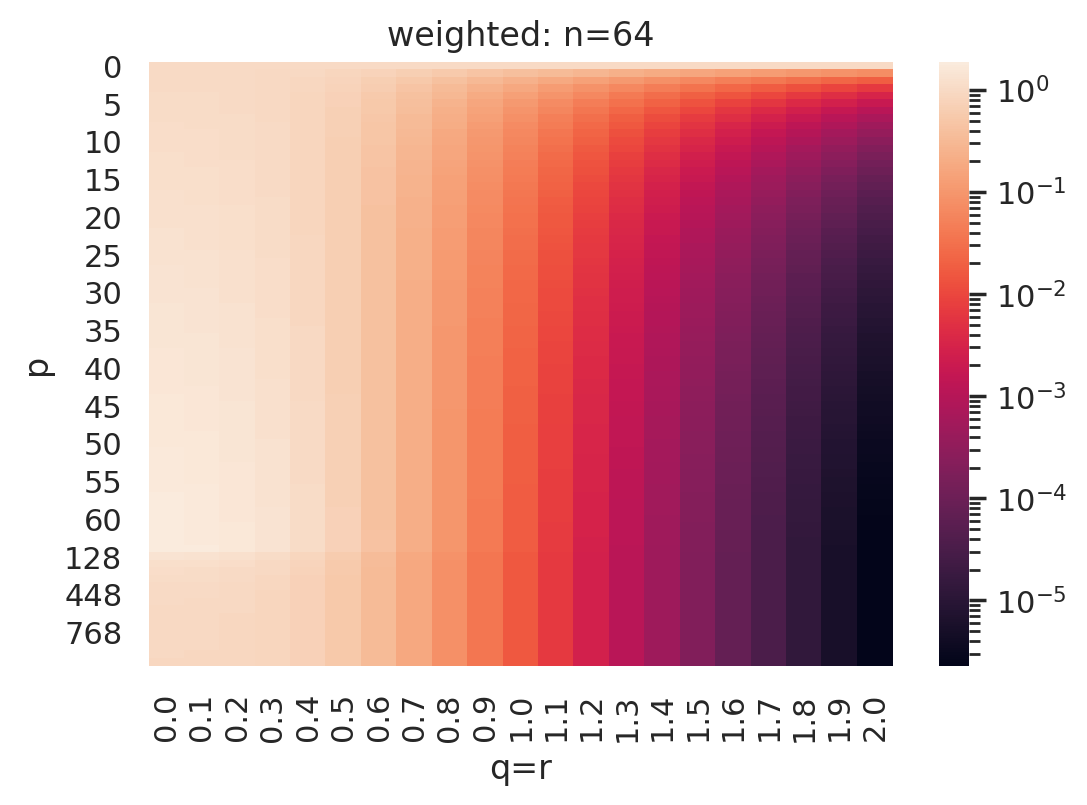}
    \includegraphics[width=.28\linewidth]{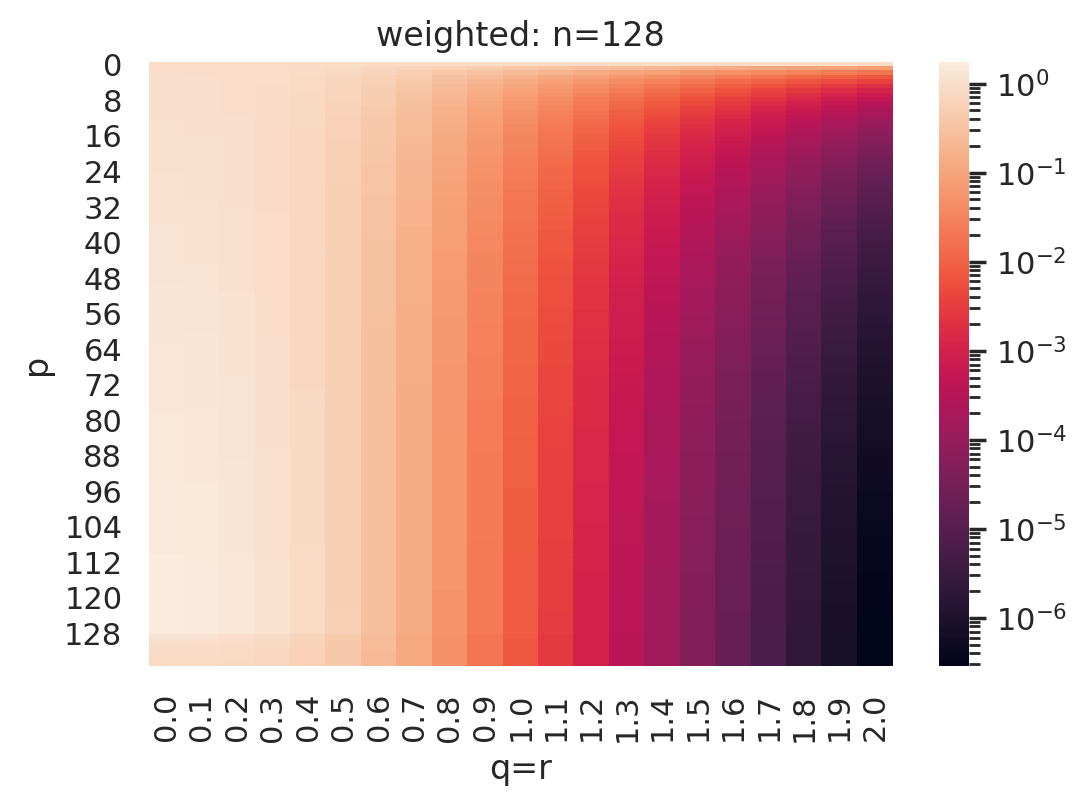}
    \includegraphics[width=.28\linewidth]{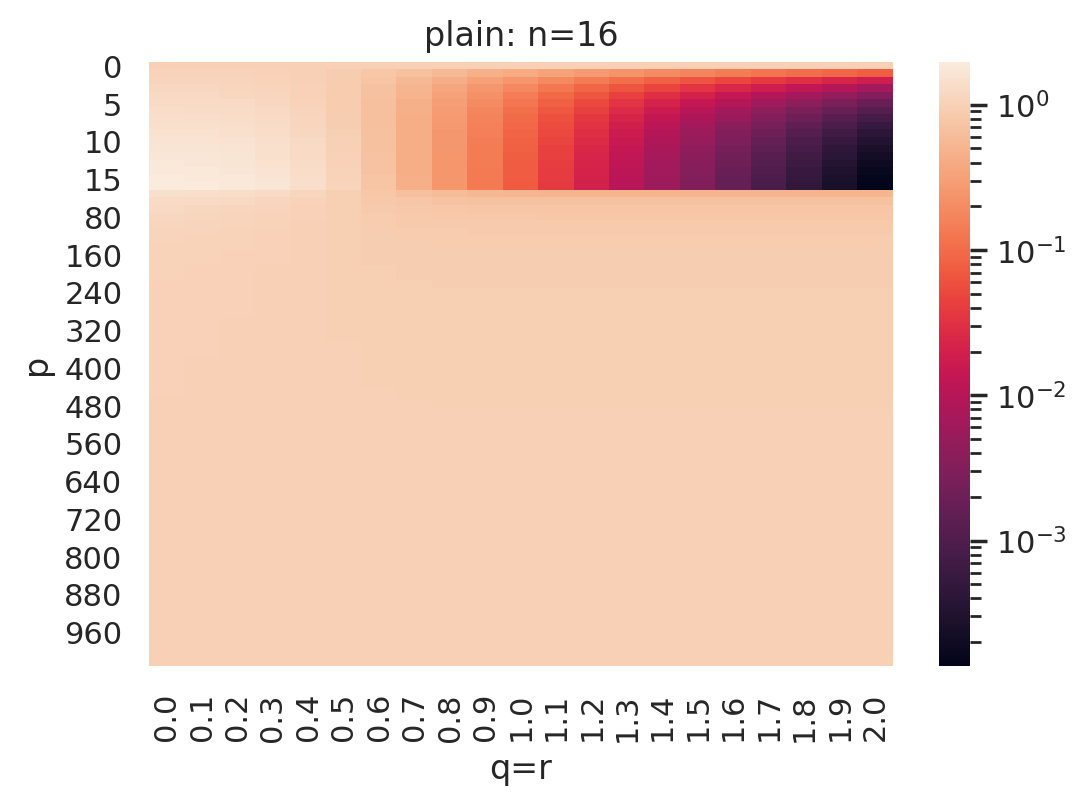}
    \includegraphics[width=.28\linewidth]{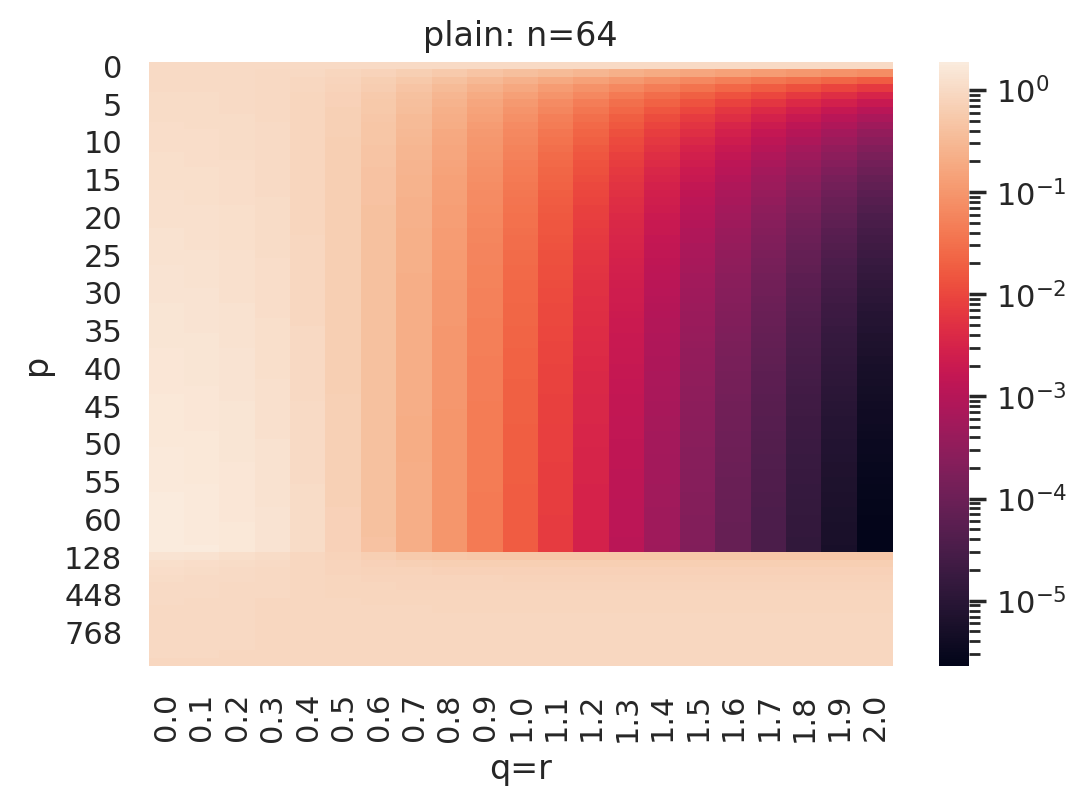}
    \includegraphics[width=.28\linewidth]{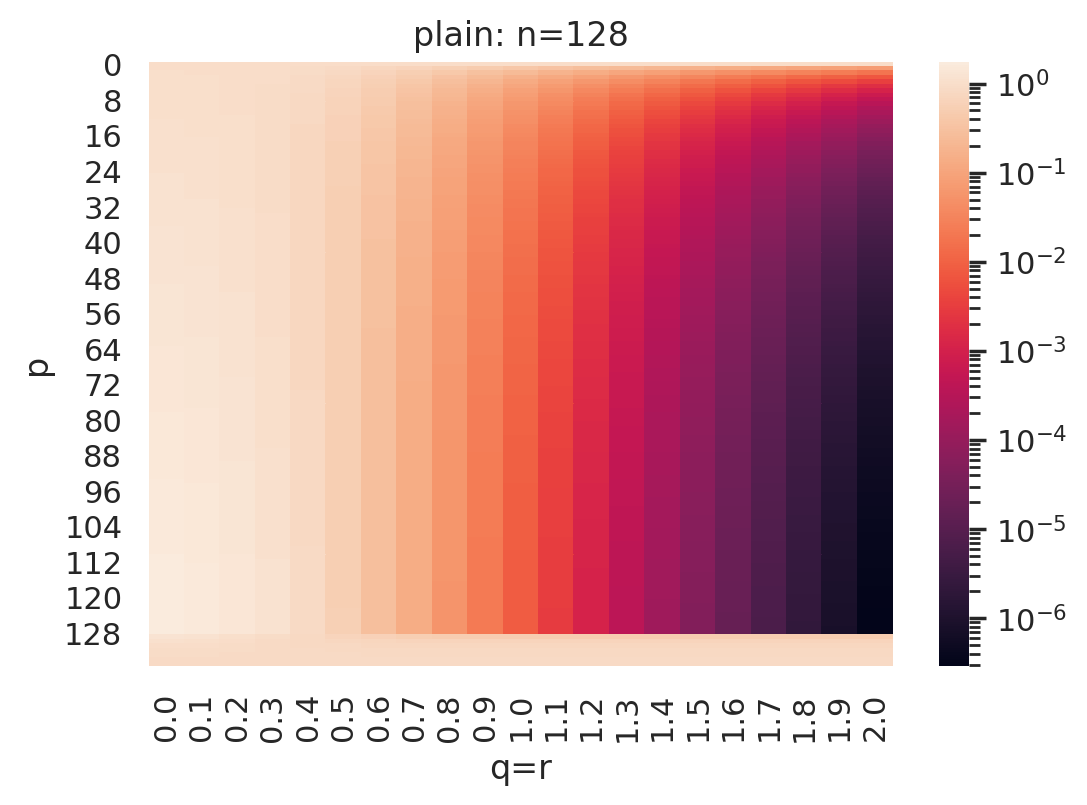}
    \vspace{-.3cm}
    \caption{Heat maps of theoretical risks of weighted (up) and plain min-norm (down) estimators for $r$-decaying features with $D=1024, q=r$, and $n=16, 64, 128$. (Note the these heat maps on the right are not corrupted: there are light color blocks since the risks of the plain min-norm estimators (i.e., $q=0$ in Figure \ref{fig:thm-lines-first}) changes to around $1$ after $p>n$, and the color bar is in log scale. This transition also occurs with the weighted estimator, where the faint horizontal line takes place ($p=n$). It corresponds to a peak in risk, as in Figure \ref{fig:thm-lines-first}.) }
    \vspace{-.5cm}
    \label{fig:heat-weight}
\end{figure}
\subsection{Heat Maps of Theoretical Risks} We show the heat maps for the theoretical risks corresponding to weighted and plain min-norm estimators in Figure \ref{fig:heat-weight}, which are calculated by Theorem \ref{thm:diff-equi} and \ref{thm:under}. Here, we use Fourier series models with $D=1024$, varying $n$, and $q=r$. The x-axis is $r$ of the $r$-decaying coefficients (from $0$ to $2$ with $0.1$ as the step), the y-axis is $p$ (where $p<n$ in the underparameterized regime and $p=l n, l\in \sN_+$ in the overparameterized regime), and the risks are in log scale. We can see the trends of the risks: the top three plots show that when $q=r>1$ the risk monotonically decreases as $p$ increases in the underparameterized regime and the lowest risk lies in the overparameterized regime; the bottom three plots show that after $p>n$, the risks for the plain min-norm estimator ($q=0$) increase suddenly, and they are higher (i.e., the light color block in each heat map) than the risks in then underparameterized regime when $r>1$. Hence, these plots also verify Theorem \ref{thm:lower bound}.
\begin{figure}[ht]
    \centering
    \vspace{-.3cm}
    \includegraphics[width=.85\linewidth]{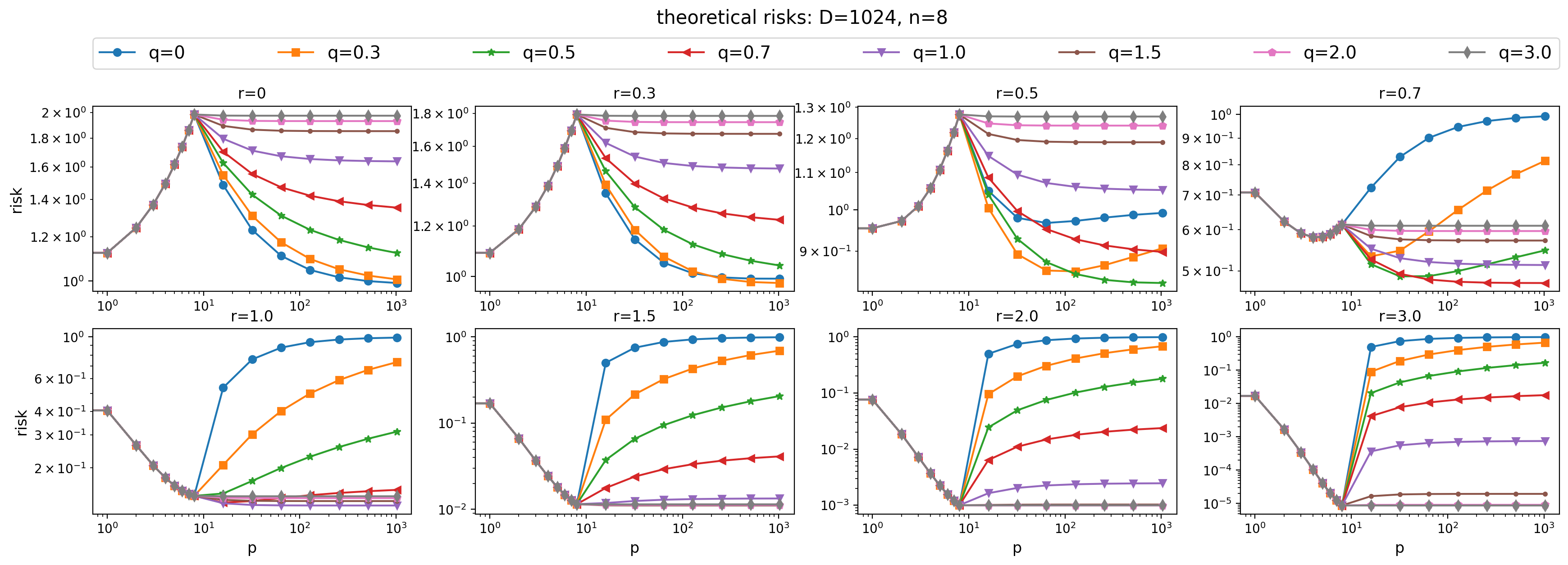}
    \includegraphics[width=.85\linewidth]{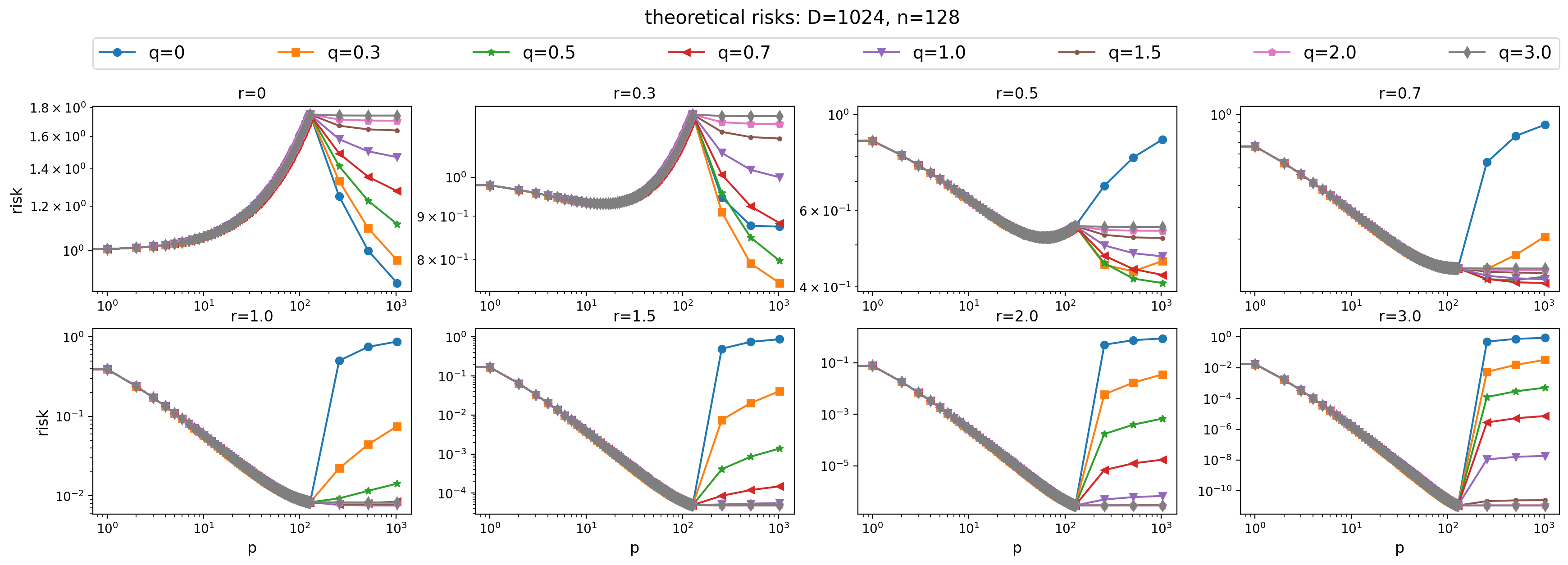}
    \vspace{-.4cm}
    \caption{Theoretical risks for $r$-decaying features and varying $q$ with $D=1024$,
    up: $n=8$, down: $n=128$.}
    \vspace{-.5cm}
    \label{fig:thm-lines-first}
\end{figure}

\subsection{Theoretical Risks with Varying r and q}
Figure~\ref{fig:thm-lines-first} shows the plots of the theoretical extended risk curves (fixed $n$) with a range of choices for $r$ and $q$ as mentioned in Section \ref{sec:underpara}, from which we can see the trends and patterns of the risks. In this experiment, we use Fourier series models with $D=1024$, $n=8$ and $128$, $p<n$ in the underparameterized regime and $p=ln, l\in \sN_+$ in the overparameterized regime. We investigate on $r=0, 0.3, 0.5, 0.7, 1.0, 1.5 ,2.0, 3.0$ and {\color{black}$q$ within the same range,} but not necessarily equal to $r$. The curves with $q = 0$ correspond to the risks for the plain min-norm estimator. Some observations of the plots are as follows.
\begin{enumerate}
    \item For varying $n$, the trends with the same $r$ and $q$ are similar along with different transition points ($p=n$), except for the case $r=0$. {\color{black}For a fixed $r$, the theoretical line with $q=r$ obtains the lowest risk among all choices of $q$, which empirically verifies our suggestion on the choice of $q$ in Remark \ref{rmk:lowrisk}. See the robustness of $q \approx r$ in Figure \ref{fig:q_approx_r}.}
    \item In the underparameterized regime, when $r=0$ and $n <D/2$, the risk increases with $p$ until $p=n$. The phase transition from $r=0$ to $r \geq 1$ validates Theorem \ref{thm:lower bound}.
    \item In the overparameterized regime, the risk of the plain min-norm estimator is almost above the weighted min-norm estimator when $r > 0.5$. Even if the 
    weight matrix does not match the covariance matrix exactly for $r$-decaying coefficients, the weighted min-norm estimator usually achieves lower risks than the plain min-norm estimator. 
    \item As stated in the proof of Theorem \ref{thm:lower bound}, the plots also show that when $q\geq r$, {\color{black} the risk in the overparameterized regime} is strictly lower than that at $p=n$, and that $r\geq 1$ is a sufficient condition to assure the monotonic decrease when $p<n$ and that the lowest risk in the over- is strictly lower than that in the under-parameterized regime.
\end{enumerate}

\subsection{Robustness of Weighted Optimization} 
\begin{figure}[ht]
    \centering
    \vspace{-.2cm}
    \includegraphics[width=\linewidth]{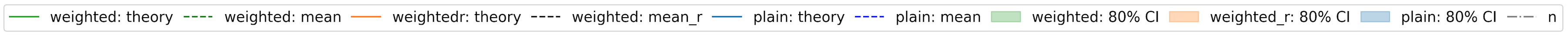}
    \includegraphics[width=.24\linewidth]{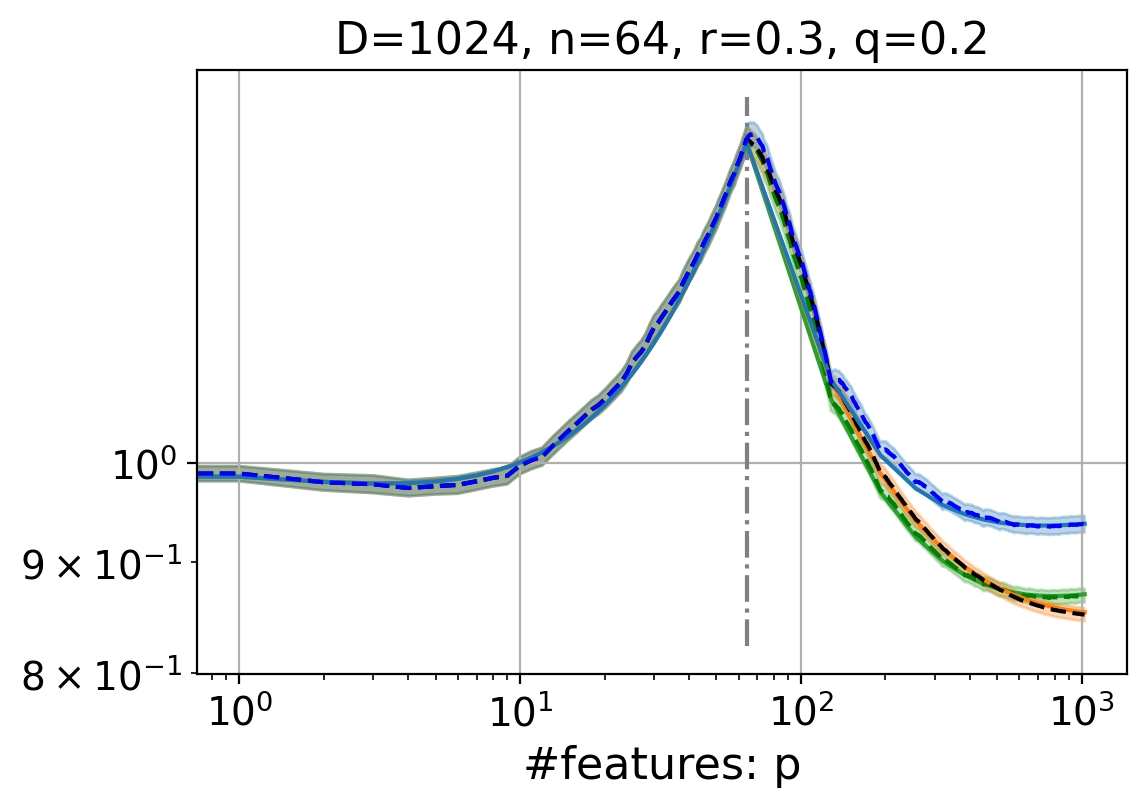}
    \includegraphics[width=.24\linewidth]{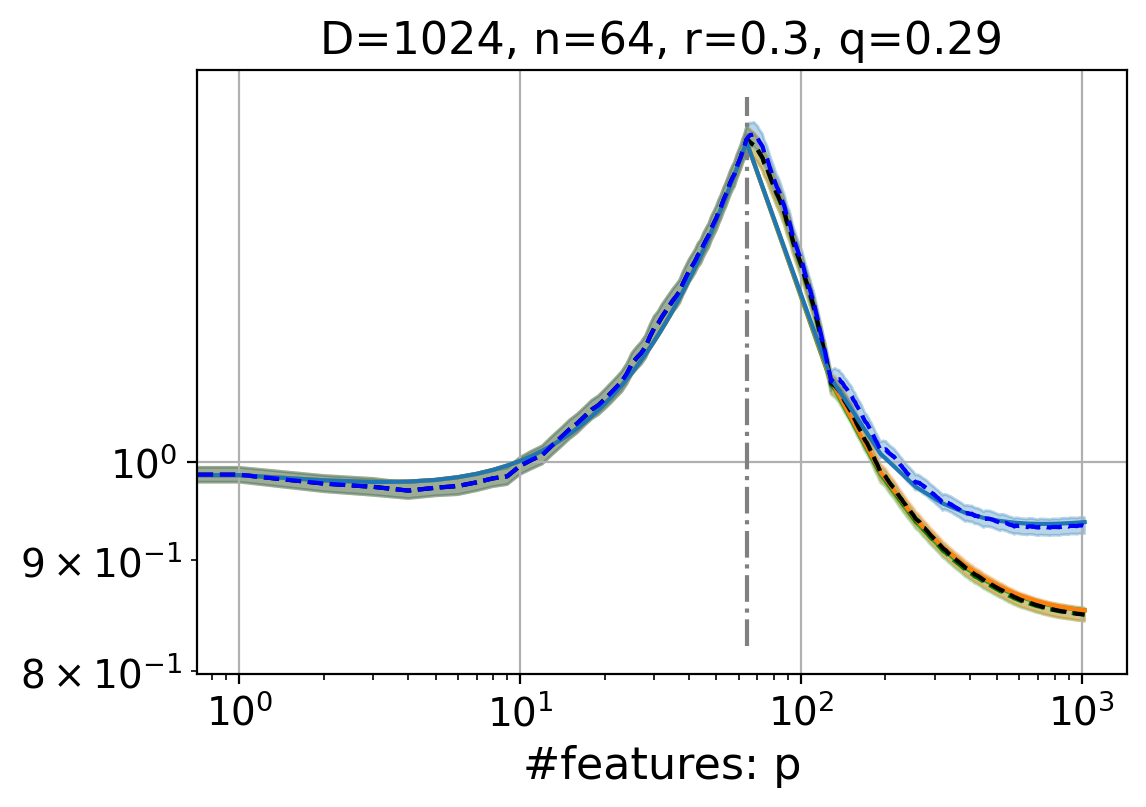}
    \includegraphics[width=.24\linewidth]{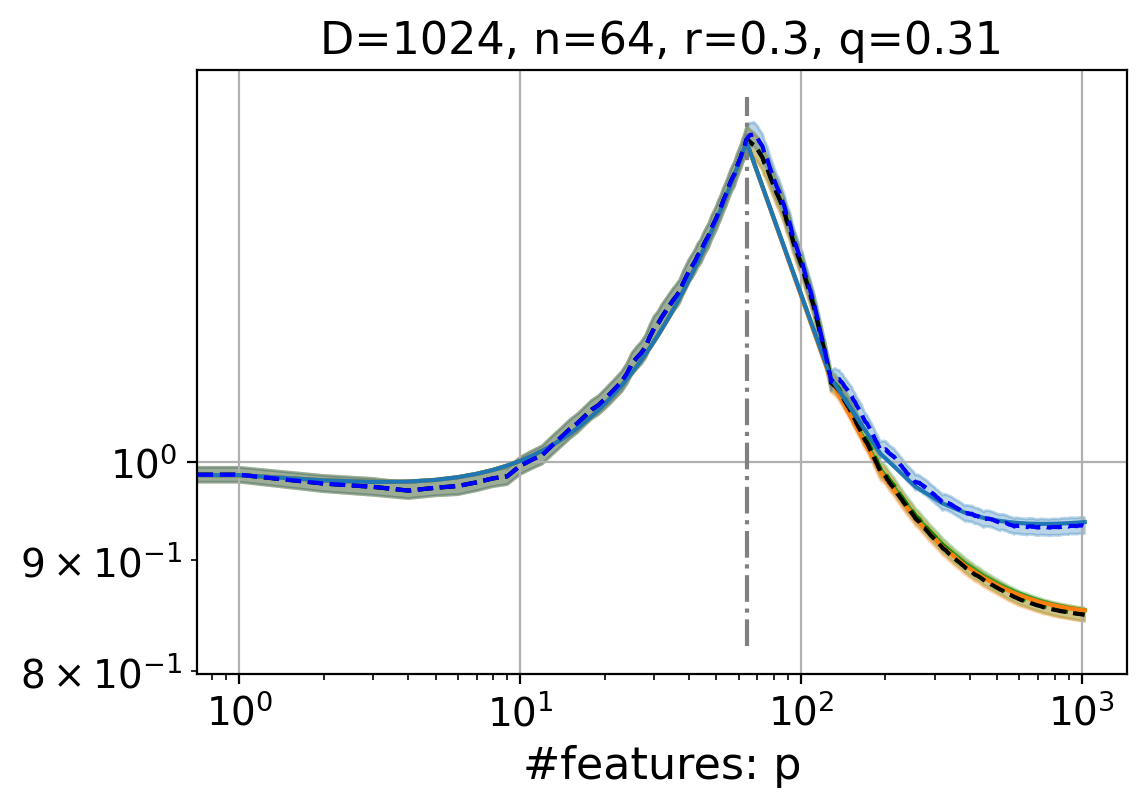}
    \includegraphics[width=.24\linewidth]{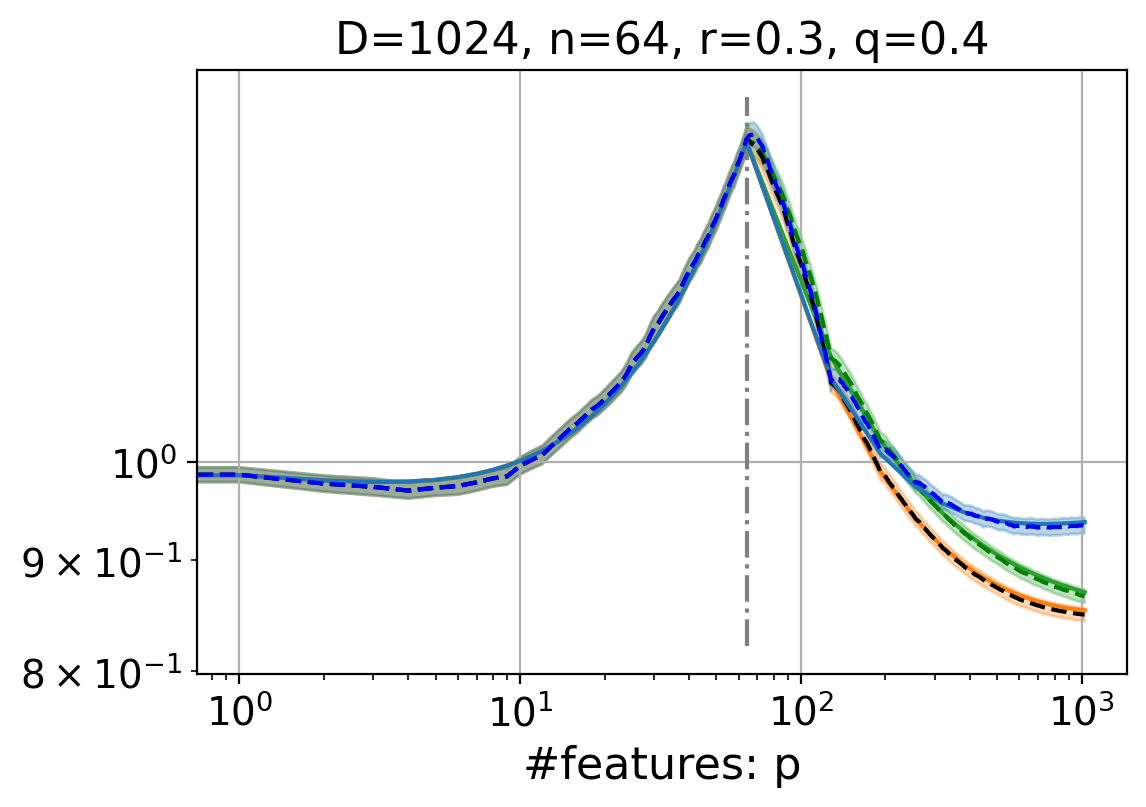}
    \includegraphics[width=.24\linewidth]{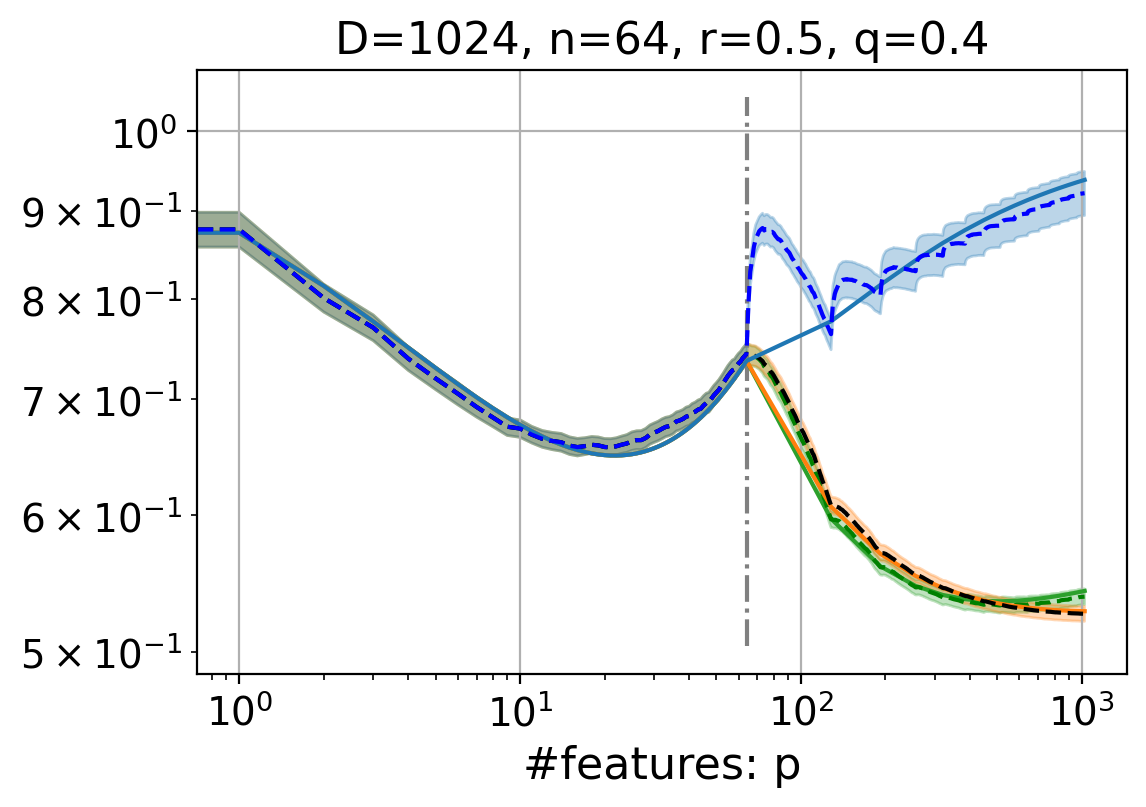}
    \includegraphics[width=.24\linewidth]{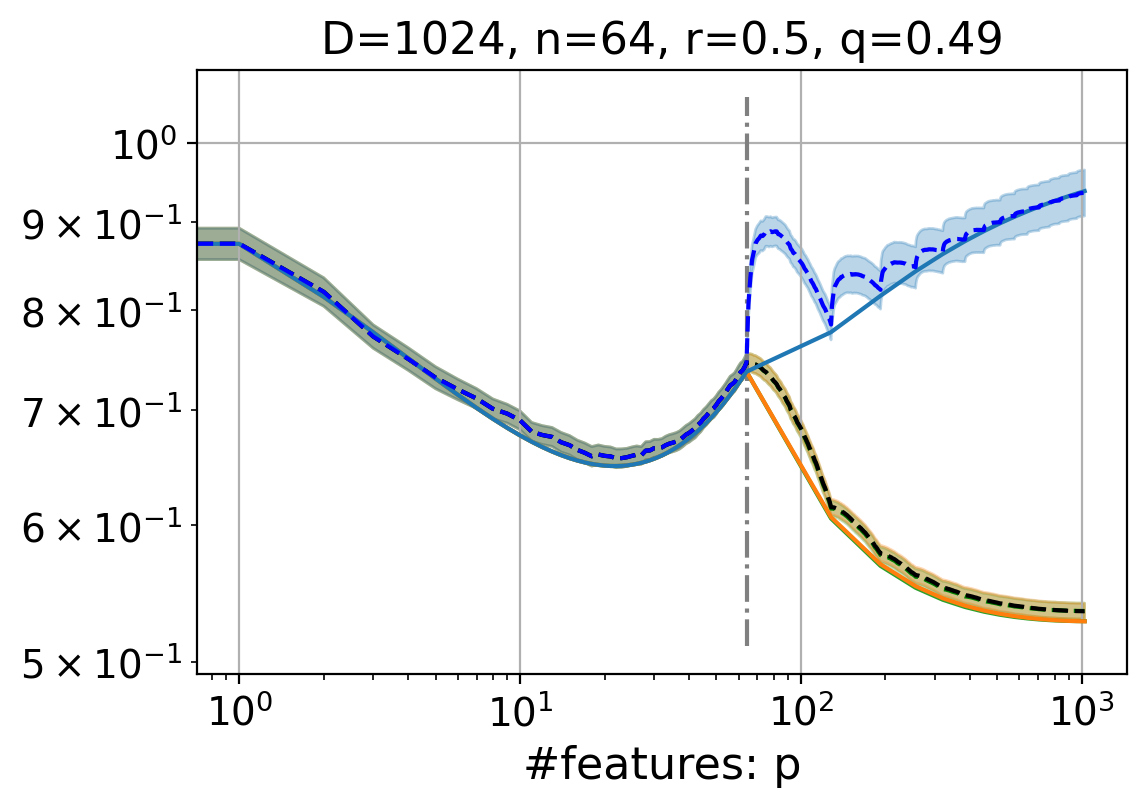}
    \includegraphics[width=.24\linewidth]{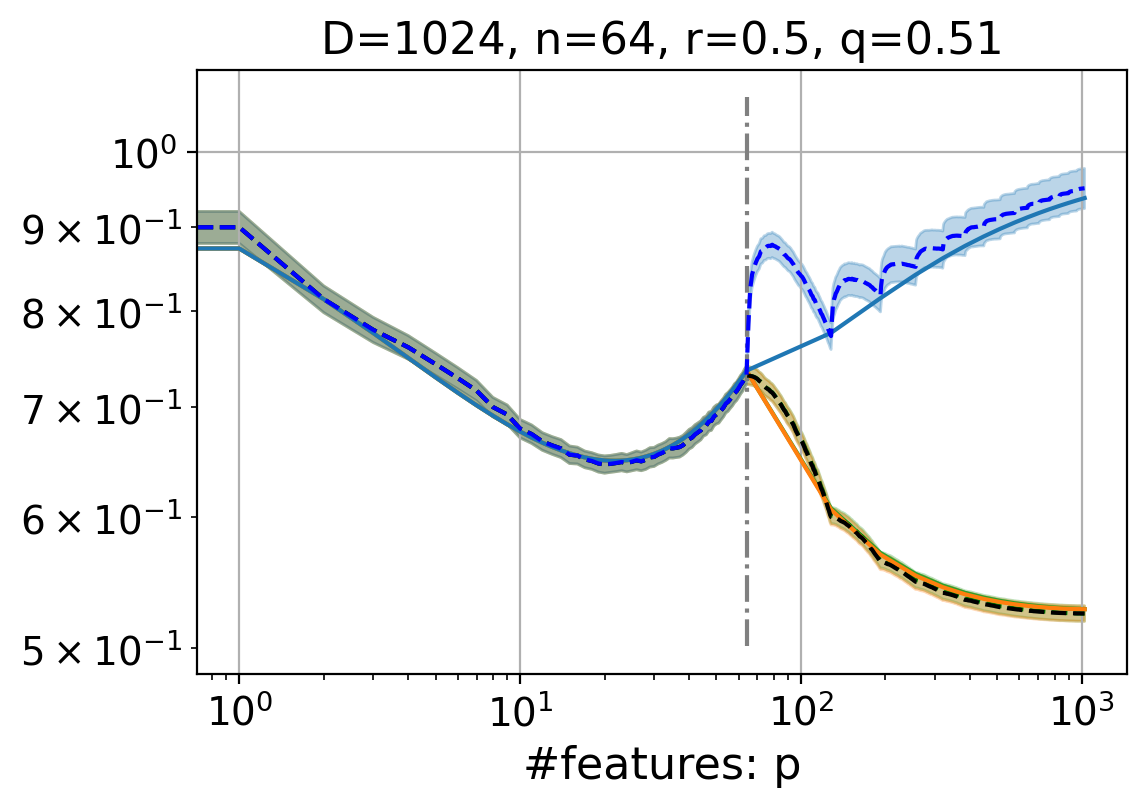}
    \includegraphics[width=.24\linewidth]{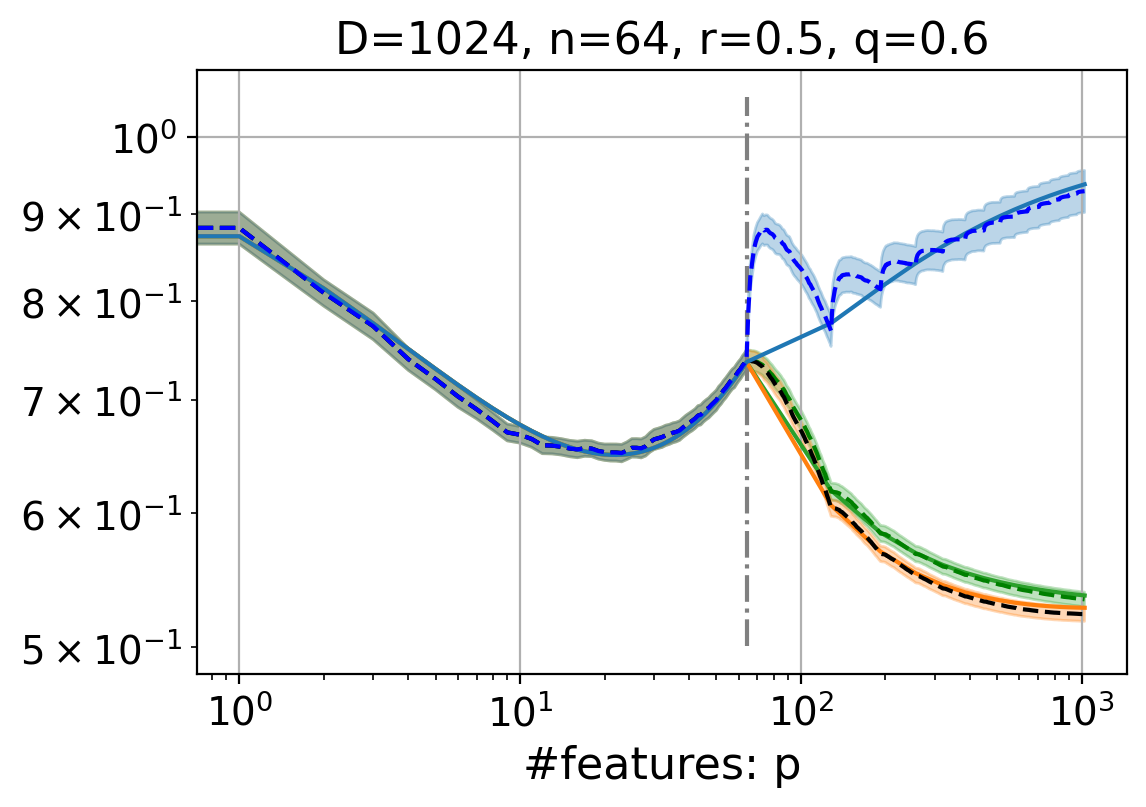}
    \includegraphics[width=.24\linewidth]{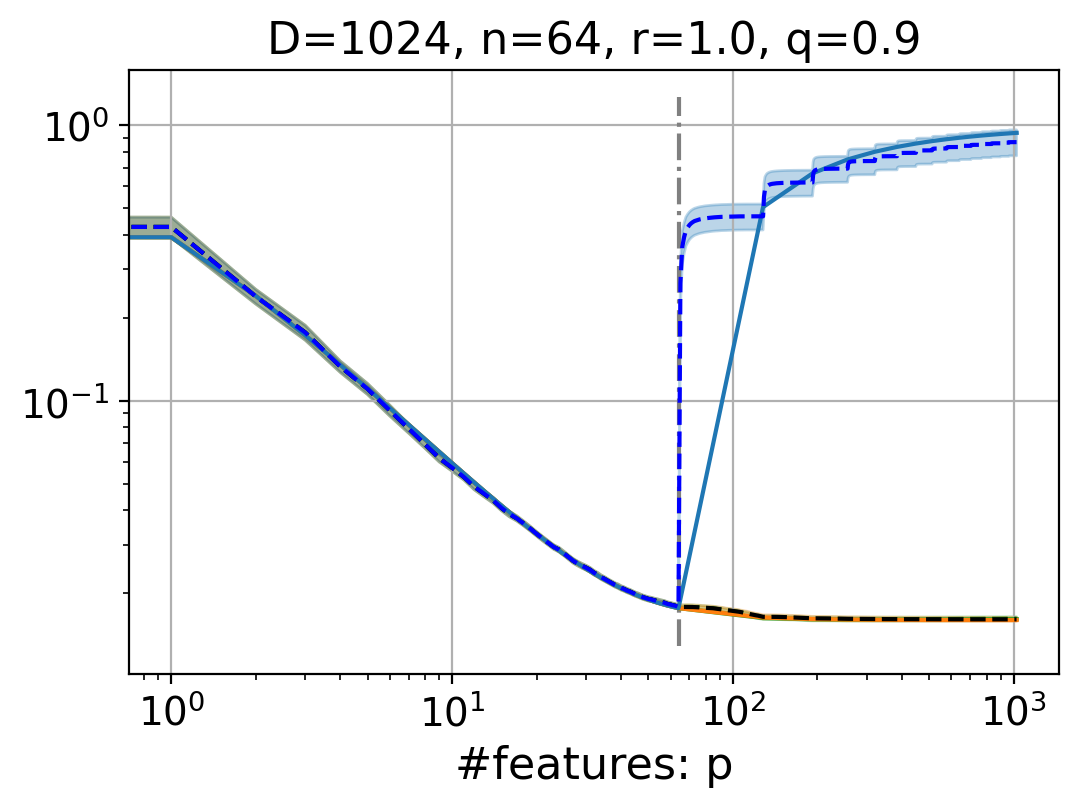}
    \includegraphics[width=.24\linewidth]{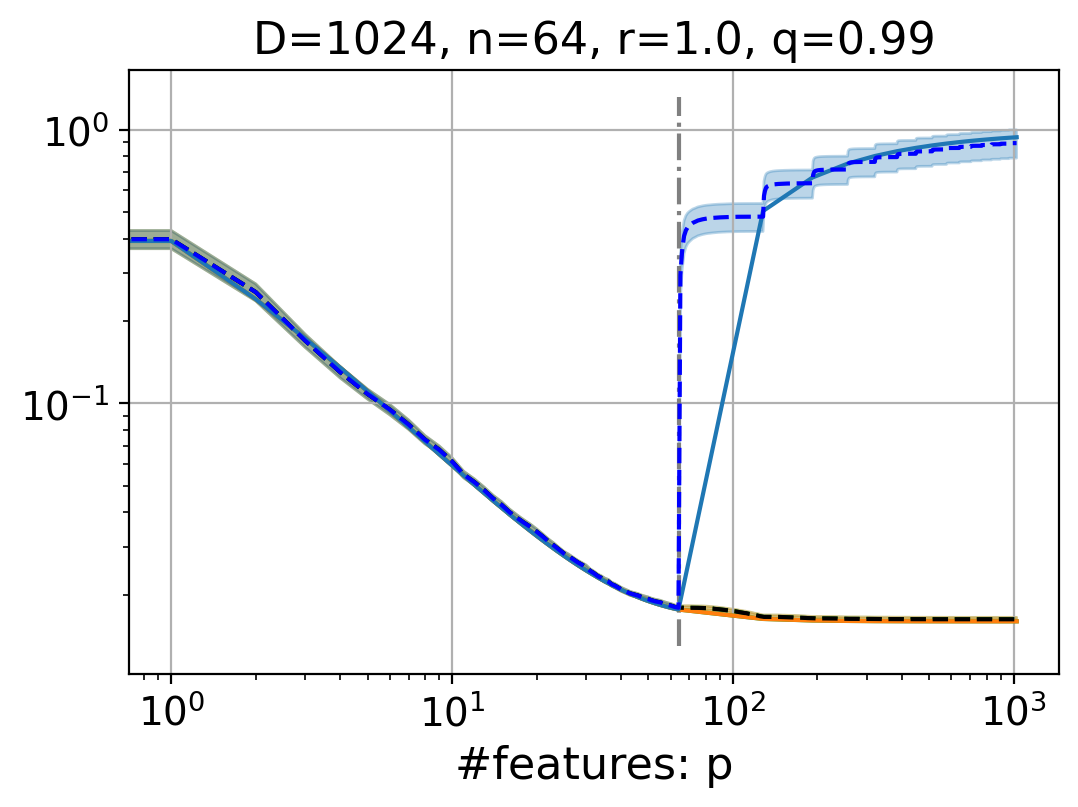}
    \includegraphics[width=.24\linewidth]{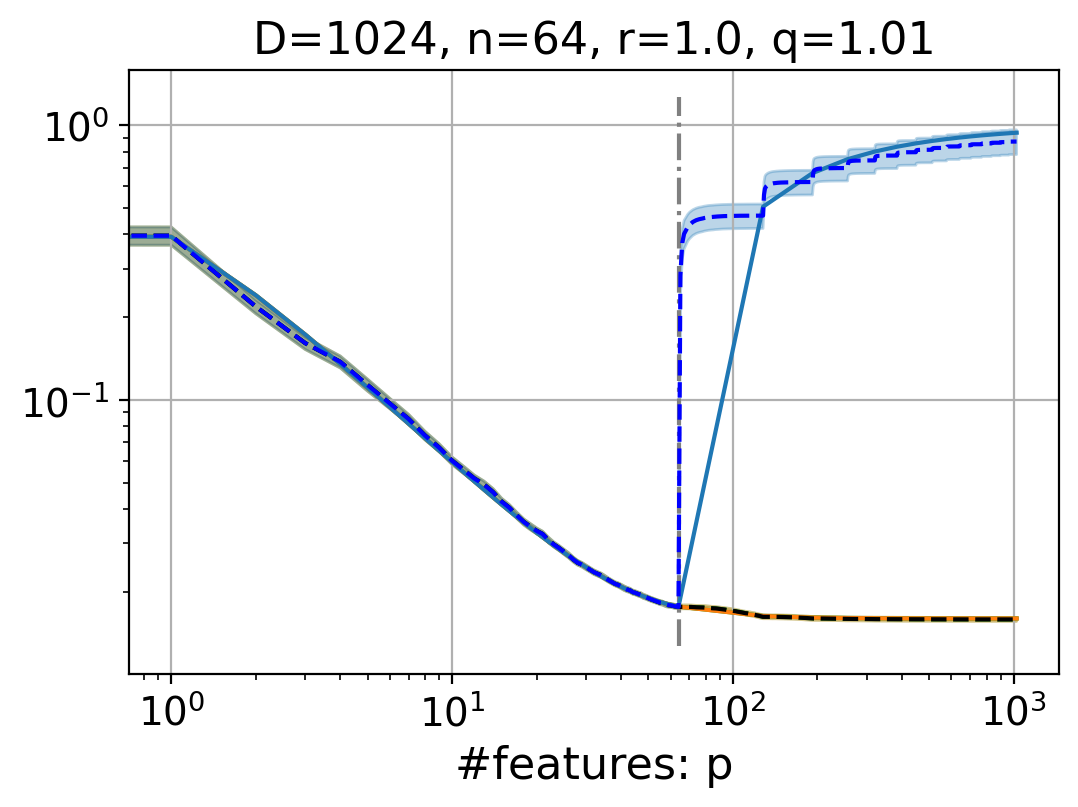}
    \includegraphics[width=.24\linewidth]{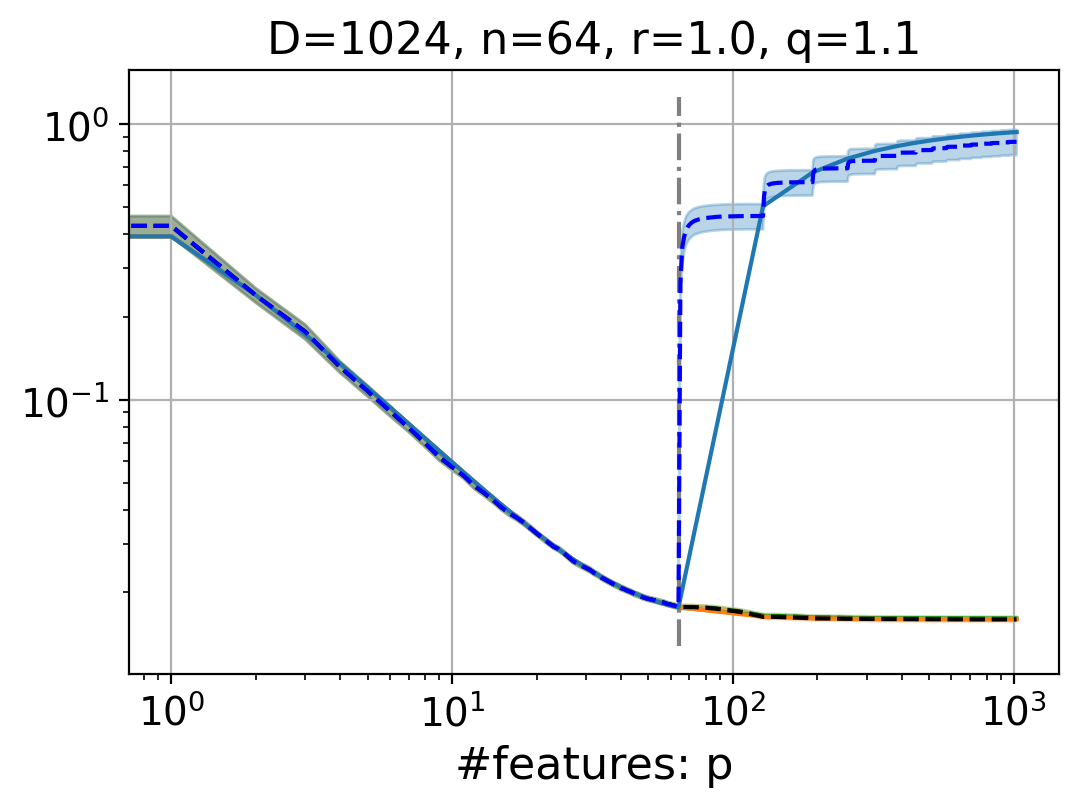}
    \vspace{-.4cm}
    \caption{Theoretical and empirical risks $(\|\vtheta - \htheta \|_2^2)$ of plain and weighted min-norm estimators with $q\approx r$. From top to bottom: $r= 0.3, 0.5, 1.0$; From left to right: $q=r-0.1, r-0.01, r+0.01, r+0.1$ (in green) with the base case $q=r$ (in orange).}
    \label{fig:q_approx_r}
    \vspace{-.3cm}
    \end{figure}

{\color{black} Figure \ref{fig:q_approx_r} shows empirical and theoretical risks for the weighted min-norm estimator when $q\approx r$ is close to that of $q=r$. The theoretical risk lines match the mean of $\|\vtheta - \htheta\|_2$ out of $100$ runs.}

\bibliographystyle{siamplain}
\bibliography{references}

\end{document}